\documentclass[10pt,twocolumn,letterpaper]{article}

\usepackage[pagenumbers]{cvpr} 

\usepackage{graphicx}
\usepackage{amsmath}
\usepackage{amssymb}
\usepackage{booktabs}
\usepackage{algorithm}
\usepackage{algpseudocode}

\usepackage{times}
\usepackage{epsfig}
\usepackage{graphicx}
\usepackage{amsmath, bm}
\usepackage{amssymb}
\usepackage{multirow}
\usepackage{microtype}
\usepackage[normalem]{ulem}
\useunder{\uline}{\ul}{}
\usepackage{color}
\usepackage{nicefrac}
\usepackage{comment}
\usepackage{enumitem}
\usepackage{adjustbox}
\usepackage{mathabx}
\usepackage{tabu}
\usepackage{booktabs}
\usepackage{diagbox}
\usepackage{cuted}
\usepackage{capt-of} 
\usepackage{multicol}
\usepackage[table,x11names]{xcolor}
\usepackage[subtle,margins=normal,leading=normal]{savetrees}

\usepackage[pagebackref,breaklinks,colorlinks]{hyperref}
\usepackage[capitalize]{cleveref}
\crefname{section}{Sec.}{Secs.}
\Crefname{section}{Section}{Sections}
\Crefname{table}{Table}{Tables}
\crefname{table}{Tab.}{Tabs.}
\usepackage{appendix}

\def\cvprPaperID{121} 
\def\confName{CVPR}
\def\confYear{2022}

\begin{document}

\title{\Large Self-Supervised Pre-Training of Swin Transformers 

for 3D Medical Image Analysis}

\author{
Yucheng Tang$^*$\\
Vanderbilt University\\
\and
Dong Yang\\
NVIDIA\\
\and
Wenqi Li\\
NVIDIA\\
\and
Holger R. Roth\\
NVIDIA\\
\and
Bennett Landman\\
Vanderbilt University\\
\and
Daguang Xu\\
NVIDIA\\
\and
Vishwesh Nath$^\dagger$\\
NVIDIA\\
\and
Ali Hatamizadeh$^{\dagger \ddagger}$\\
NVIDIA\\
}

\maketitle
\def\thefootnote{$*$}\footnotetext{Work completed during internship at NVIDIA}\def\thefootnote{\arabic{footnote}}
\def\thefootnote{$\dagger$}\footnotetext{Co-senior advising}\def\thefootnote{\arabic{footnote}}
\def\thefootnote{$\ddagger$}\footnotetext{Corresponding author: \texttt{ahatamizadeh@nvidia.com} }\def\thefootnote{\arabic{footnote}}

\begin{abstract}
   Vision Transformers (ViT)s have shown great performance in self-supervised learning of global and local representations that can be transferred to downstream applications. Inspired by these results, we introduce a novel self-supervised learning framework with tailored proxy tasks for medical image analysis. Specifically, we propose: (i) a new 3D transformer-based model, dubbed Swin UNEt TRansformers (Swin UNETR), with a hierarchical encoder for self-supervised pre-training; (ii) tailored proxy tasks for learning the underlying pattern of human anatomy. We demonstrate successful pre-training of the proposed model on 5,050 publicly available computed tomography (CT) images from various body organs. The effectiveness of our approach is validated by fine-tuning the pre-trained models on the Beyond the Cranial Vault (BTCV) Segmentation Challenge with $13$ abdominal organs and segmentation tasks from the Medical Segmentation Decathlon (MSD) dataset. Our model is currently the state-of-the-art on the public test leaderboards of both MSD\footnote{{\url{https://decathlon-10.grand-challenge.org/evaluation/challenge/leaderboard/}}} and BTCV \footnote{{\url{https://www.synapse.org/\#!Synapse:syn3193805/wiki/217785/}}} datasets. Code: \href{https://monai.io/research/swin-unetr}{https://monai.io/research/swin-unetr}.
\end{abstract}

\section{Introduction}
\label{sec:intro}
\begin{figure}[t]
\includegraphics[width=\linewidth]{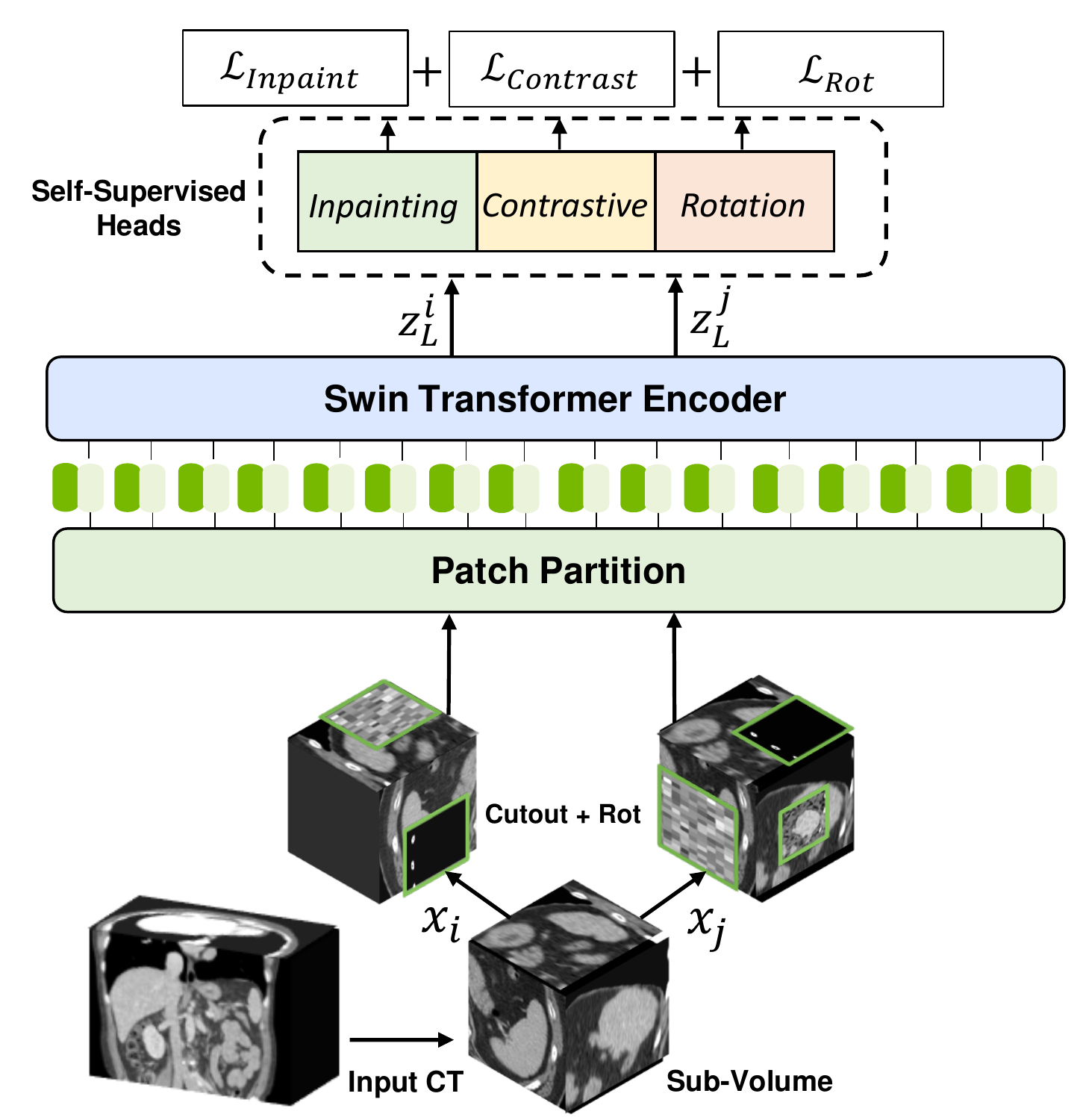}
\caption{Overview of our proposed pre-training framework. Input CT images are randomly cropped into sub-volumes and augmented with random inner cutout and rotation, then fed to the Swin UNETR encoder as input. We use masked volume inpainting, contrastive learning and rotation prediction as proxy tasks for learning contextual representations of input images.}
\label{fig:fig1}
\end{figure}
Vision Transformers (ViT)s~\cite{dosovitskiy2020image} have started a revolutionary trend in computer vision~\cite{zheng2021rethinking,cheng2021per} and medical image analysis~\cite{chen2021transunet,hatamizadeh2021unetr}. Transformers demonstrate exceptional capability in learning pre-text tasks, are effective in learning of global and local information across layers, and provide scalability for large-scale training~\cite{zhai2021scaling,raghu2021vision}. As opposed to convolutional neural networks (CNNs) with limited receptive fields, ViTs encode visual representations from a sequence of patches and leverage self-attention blocks for modeling long-range global information~\cite{raghu2021vision}. Recently, Shifted windows (Swin) Transformers~\cite{liu2021swin} proposed a hierarchical ViT that allows for local computing of self-attention with non-overlapping windows. This architecture achieves linear complexity as opposed to quadratic complexity of self-attention layers in ViT, hence making it more efficient. In addition, due to the hierarchical nature of Swin Transformers, they are well-suited for tasks requiring multi-scale modeling.

In comparison to CNN-based counterparts, transformer-based models learn stronger features representations during pre-training, and as a result perform favorably on fine-tuning downstream tasks~\cite{raghu2021vision}. Several recent efforts on ViTs~\cite{caron2021emerging,xie2021self} have achieved new state-of-the-art results by self-supervised pre-training on large-scale datasets such as ImageNet~\cite{deng2009imagenet}. 

In addition, medical image analysis has not benefited from these advances in general computer vision due to: (1) large domain gap between natural images and medical imaging modalities, like computed tomography (CT) and magnetic resonance imaging (MRI); (2) lack of cross-plane contextual information when applied to volumetric (3D) images (such as CT or MRI). The latter is a limitation of 2D transformer models for various medical imaging tasks such as segmentation. Prior studies have demonstrated the effectiveness of supervised pre-training in medical imaging for different applications~\cite{chen2019med3d, raghu2019transfusion}. But creating expert-annotated 3D medical datasets at scale is a non-trivial and time-consuming effort. 

To tackle these limitations, we propose a novel self-supervised learning framework for 3D medical image analysis. First, we propose a new architecture dubbed Swin UNEt TRansformers (Swin UNETR) with a Swin Transformer encoder that directly utilizes 3D input patches. Subsequently, the transformer encoder is pre-trained with tailored, self-supervised tasks by leveraging various proxy tasks such as image inpainting, 3D rotation prediction, and contrastive learning (See Fig.~\ref{fig:fig1} for an overview). Specifically, the human body presents naturally consistent contextual information in radiographic images such as CT due to its depicted anatomical structure~\cite{yan2020self, tang2021body}. Hence, proxy tasks are utilized for learning the underlying patterns of the human anatomy. For this purpose, we extracted numerous patch queries from different body compositions such as head, neck, lung, abdomen, and pelvis to learn robust feature representations from various anatomical contexts, organs, tissues, and shapes. 

Our framework utilizes contrastive learning~\cite{oord2018representation}, masked volume inpainting~\cite{pathak2016context}, and 3D rotation prediction~\cite{gidaris2018unsupervised} as pre-training proxy tasks. The contrastive learning is used to differentiate various ROIs of different body compositions, whereas the inpainting allows for learning the texture, structure and correspondence of masked regions to their surrounding context. The rotation task serves as a mechanism to learn the structural content of images and generates various sub-volumes that can be used for contrastive learning. We utilize these proxy tasks to pre-train our proposed framework on a collection of $5,050$ CT images that are acquired from various publicly available datasets.

Furthermore, to validate the effectiveness of pre-training, we use 3D medical image segmentation as a downstream application and reformulate it as a 1D sequence-to-sequence prediction task. For this purpose, we leverage the Swin UNETR encoder with hierarchical feature encoding and shifted windows to extract feature representations at four different resolutions. The extracted representations are then connected to a CNN-based decoder. A segmentation head is attached at the end of the decoder for computing the final segmentation output. We fine-tune Swin UNETR with pre-trained weights on two publicly available benchmarks of Medical Segmentation Decathlon (MSD) and the Beyond the Cranial Vault (BTCV). Our model is currently the state-of-the-art on their respective public test leaderboards.  

Our main contributions in this work are summarized as follows: 
\begin{itemize}
\item We introduce a novel self-supervised learning framework with tailored proxy tasks for pre-training on CT image datasets. To this end, we propose a novel 3D transformer-based architecture, dubbed as Swin UNETR, consisting of an encoder that extracts feature representations at multiple resolutions and is utilized for pre-training.
\item We demonstrate successful pre-training on a cohort of 5,050 publicly available CT images from various applications using the proposed encoder and proxy tasks. This results in a powerful pre-trained model with robust feature representation that could be utilized for various medical image analysis downstream tasks.   
\item We validate the effectiveness of proposed framework by fine-tuning the pre-trained Swin UNETR on two public benchmarks of MSD and BTCV and achieve \emph{\textbf{state-of-the-art}} on the test leaderboards of both datasets.
\end{itemize}

\begin{figure*}[t]
\centering
\includegraphics[width=\textwidth]{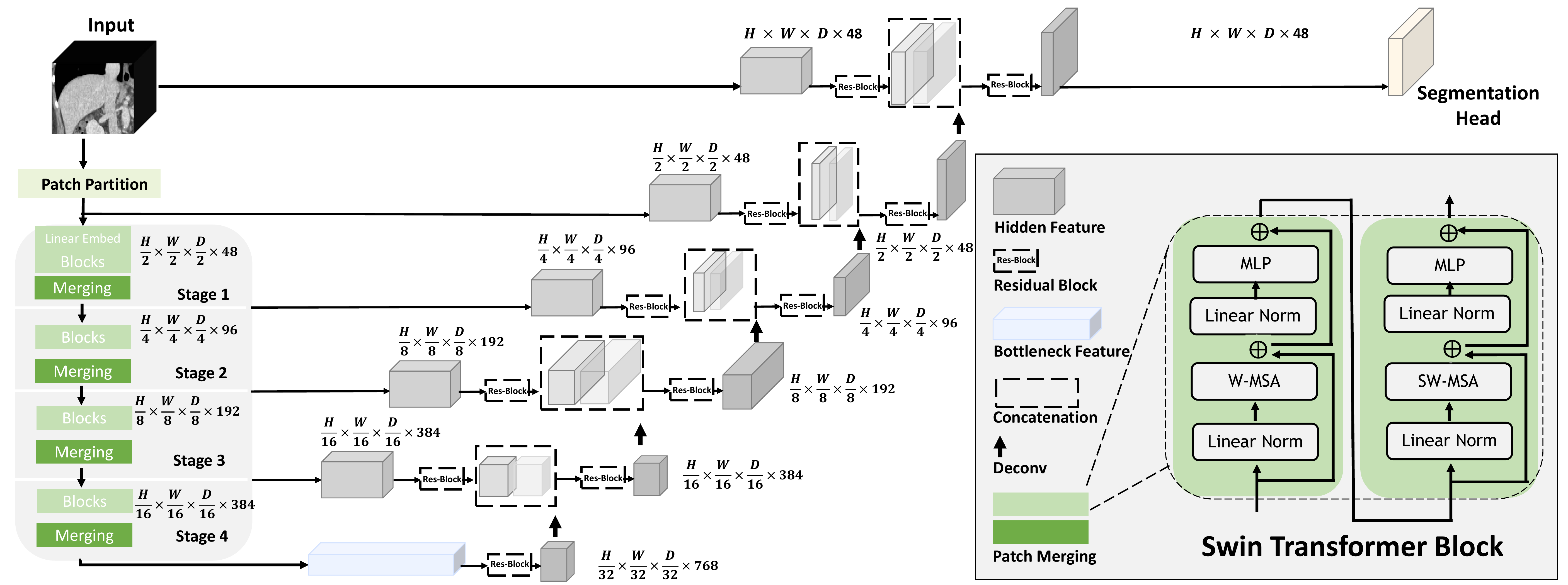}
  \caption{Overview of the Swin UNETR architecture.}
  \label{fig:fig2}
\end{figure*}
\section{Related Works}
\label{sec:formatting}

\paragraph{Medical Segmentation with Transformers}
Vision transformers are first used in classification tasks and are adopted from sequence-to-sequence modeling in natural language processing. Self-attention mechanisms that aggregate information from the entire input sequence are first achieving comparable, then better performance against prior arts of convolutional architectures such as ResNet~\cite{he2016deep} or U-Net~\cite{cciccek20163d}. Recently, transformer-based networks~\cite{xie2021cotr, zhou2021nnformer, jose2021medical, xu2021levit} are proposed for medical image segmentation. In these pioneering works, the transformer blocks are used either as a bottleneck feature encoder or as additional modules after convolutional layers, resulting in limited exploitation of the spatial context advantages of transformers. Comparing to prior works~\cite{chen2021transunet, xie2021cotr}, which are using transformers as secondary encoder, we propose to utilize transformers to embed high-dimensional volumetric medical images, which allow for a more direct encoding of 3D patches and positional embeddings. 

Most medical image analysis tasks such as segmentation requires dense inference from multi-scale features. Skip connection-based architectures such as UNet~\cite{cciccek20163d} and pyramid networks~\cite{roth2018multi} are widely adopted to leverage hierarchical features. However, vision transformers with a single patch size, while successful in natural image applications, are intractable for high-resolution and high-dimensional volumetric images. To avoid quadratic overflow of computing self-attention at scales~\cite{lin2017feature, singh2018sniper}, Swin Transformer~\cite{liu2021swin,liu2021video} is proposed to construct hierarchical encoding by a shifted-window mechanism. Recent works such as Swin UNet~\cite{cao2021swin} and DS-TransUNet~\cite{lin2021ds} utilize the merits of Swin Transformers for 2D segmentation and achieve promising performance. Augmenting the above-mentioned methods, we learn from 3D anatomy in broader  medical image segmentation scenarios by incorporating hierarchically volumetric context.
\paragraph{Pre-training in Medical Image Analysis}
In medical image analysis, previous studies of pre-training on labeled data demonstrate improved performance by transfer learning~\cite{chen2019med3d, raghu2019transfusion}. However, generating annotation for medical images is expensive and time-consuming. Recent advances in self-supervised learning offer the promise of utilizing unlabeled data. Self-supervised representation learning~\cite{atito2021sit, dai2021up, liang2021swinir} constructs feature embedding spaces by designing pre-text tasks, such as solving jigsaw puzzles~\cite{noroozi2016unsupervised}. Another commonly used pre-text task is to memorize spatial context from medical images, which is motivated by image restoration. This idea is generalized to inpainting tasks~\cite{zhou2021models, haghighi2021transferable, pathak2016context} to learn visual representations~\cite{chen2019self, wang2021self, azizi2021big} by predicting the original image patches. Similar efforts for reconstructing spatial context have been formulated as solving Rubik's cube problem~\cite{zhu2020rubik}, random rotation prediction~\cite{gidaris2018unsupervised, NEURIPS2020_d2dc6368} and contrastive coding~\cite{chen2021empirical,oord2018representation}. Different from these efforts, our pre-training framework is simultaneously trained with a combination of pre-text tasks, tailored for 3D medical imaging data, and leverages a transformer-based encoder as a powerful feature extractor.     

\section{Swin UNETR}
Swin UNETR comprises a Swin Transformer~\cite{liu2021swin} encoder that directly utilizes 3D patches and is connected to a CNN-based decoder via skip connections at different resolutions. Fig.~\ref{fig:fig2} illustrates the overall architecture of Swin UNETR. We describe the details of encoder and decoder in this section.   
\subsection{Swin Transformer Encoder}
\begin{figure}[t]
\includegraphics[width=\linewidth]{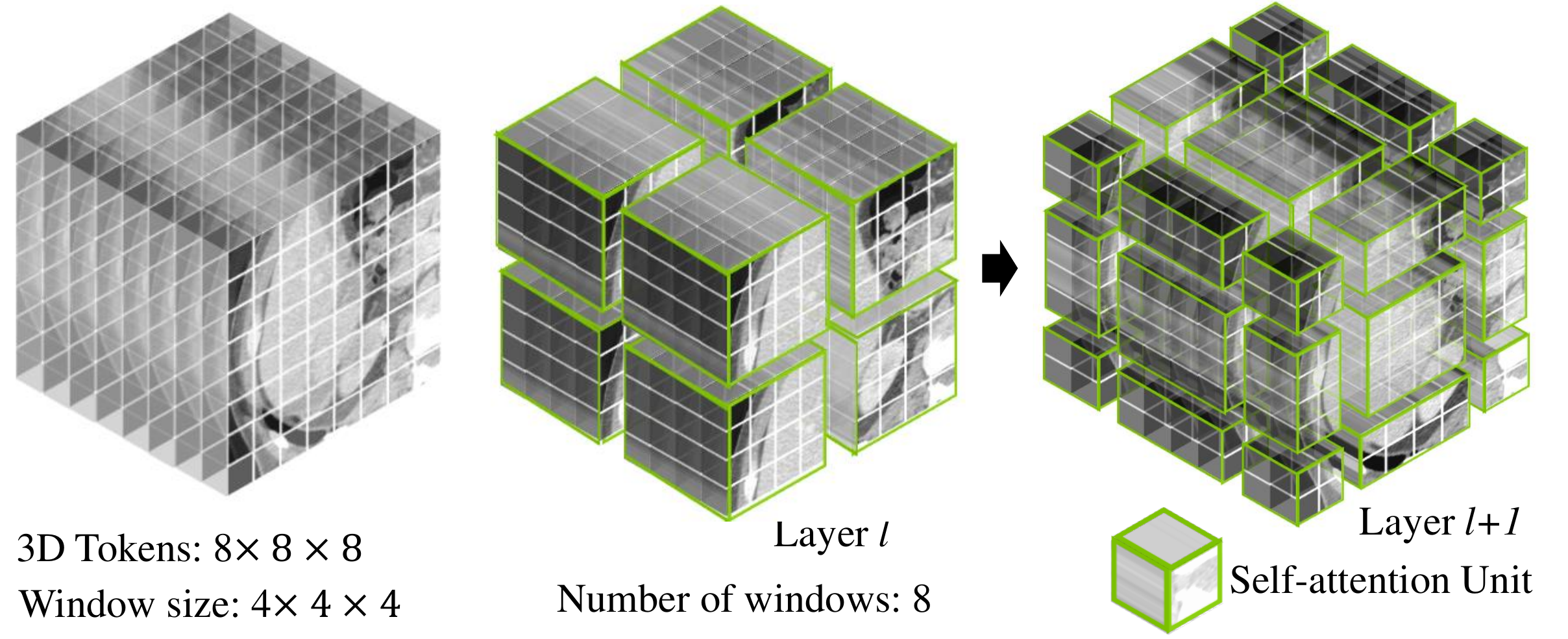}
  \caption{Shifted windowing mechanism for efficient self-attention computation of 3D tokens with $8 \times 8 \times 8$ tokens and $4 \times 4 \times 4$ window size.}
  \label{fig:fig3}
\end{figure}

Assuming that the input to the encoder is a sub-volume $\mathcal{X} \in \mathbb{R}^{H\times{W}\times{D}\times{S}}$, a 3D token with a patch resolution of $(H^{\prime},W^{\prime},D^{\prime})$ has a dimension of $H^{\prime} \times W^{\prime}\times D^{\prime}\times S$. The patch partitioning layer creates a sequence of 3D tokens with size  $\frac{H}{H^{\prime}}\times\frac{W}{W^{\prime}}\times\frac{D}{D^{\prime}}$ that are projected into a $C$-dimensional space via an embedding layer. Following~\cite{liu2021swin}, for efficient modeling of token interactions, we partition the input volumes into non-overlapping windows and compute local self-attention within each region. Specifically, at layer $l$, we use a window of size $M\times M\times M$ to evenly divide a 3D token into $\left\lceil\frac{H^{\prime}}{M}\right\rceil\times\left\lceil\frac{W^{\prime}}{M}\right\rceil \times\left\lceil\frac{D^{\prime}}{M}\right\rceil$ windows. In the subsequent layer $l+1$, we shift the partitioned windows by $\left(\left\lfloor\frac{M}{2}\right\rfloor,\left\lfloor\frac{M}{2}\right\rfloor,\left\lfloor\frac{M}{2}\right\rfloor\right)$ voxels. The shifted windowing mechanism is illustrated in Fig.~\ref{fig:fig3}. The outputs of encoder blocks in layers $l$ and $l+1$ are computed as in
\begin{equation}
\begin{array}{l}
\hat{{z}}^{l}=\text{W-MSA}(\text{LN}({z}^{l-1}))+{z}^{l-1} \\
{z}^{l}=\text{MLP}(\text{LN}(\hat{{z}}^{l}))+\hat{{z}}^{l} \\
\hat{{z}}^{l+1}=\text{SW-MSA}(\text{LN}({z}^{l}))+{z}^{l} \\
{z}^{l+1}=\text{MLP}(\text{LN}(\hat{{z}}^{l+1}))+\hat{{z}}^{l+1},
\end{array}
\label{eq:eq1}
\end{equation}
where $\text{W-MSA}$ and $\text{SW-MSA}$ denote regular and window partitioning multi-head self-attention modules, respectively,  $\hat{{z}}^{l}$ and $\hat{{z}}^{l}$ are the outputs of $\text{W-MSA}$ and $\text{SW-MSA}$; $\text{LN}$ and $\text{MLP}$ denote layer normalization and Multi-Layer Perceptron (see Fig.~\ref{fig:fig2}). Following~\cite{liu2021swin}, we adopt a 3D cyclic-shifting for efficient batch computation of shifted windowing. Furthermore, we calculate the self-attention according to
\begin{equation}
    \textnormal{Attention}(Q, K, V) = \textnormal{Softmax}\left(\frac{QK^{\top}}{\sqrt{d}}\right)V.
\label{eq:eq2}
\end{equation}
where $Q,K,V$ represent queries, keys and values respectively, $d$ is the size of the query and key.

Our encoder uses a patch size of $2 \times 2 \times 2$~with a feature dimension of $2\times2\times2\times1 =8$ (\textit{i.e.} single input channel CT images) and a $C=48$-dimensional embedding space. Furthermore, the overall architecture of the encoder consists of $4$ stages comprising of $2$ transformer blocks at each stage (\textit{i.e.} $L=8$ total layers). In between every stage, a patch merging layer is used to reduce the resolution by a factor of $2$. Stage $1$ consists of a linear embedding layer and transformer blocks that maintain the number of tokens as $\frac{H}{2} \times \frac{W}{2} \times \frac{D}{2}$. Furthermore, a patch merging layer groups patches with resolution $2 \times 2 \times 2$ and concatenates them, resulting in a $4C$-dimensional feature embedding. A linear layer is then used to downsample the resolution by reducing the dimension to $2C$. The same procedure continues in stage 2, stage 3 and stage 4 with resolutions of $\frac{H}{4} \times \frac{W}{4} \times \frac{D}{4}$, $\frac{H}{8} \times \frac{W}{8} \times \frac{D}{8}$ and $\frac{H}{16} \times \frac{W}{16} \times \frac{D}{16}$ respectively. The hierarchical representations of the encoder at different stages are used in downstream applications such as segmentation for multi-scale feature extraction.

\subsection{Decoder}
The encoder of Swin UNETR is connected to a CNN-based decoder at each resolution via skip connections to create a ``U-shaped'' network for downstream applications such as segmentation. Specifically, we extract the output sequence representations of each stage ${i}$ ($i \in \{0,1,2,3,4\})$ in the encoder as well as the bottleneck ($i=5$) and reshape them into features with size  $\frac{H}{2^{i}} \times \frac{W}{2^{i}} \times \frac{D}{2^{i}}$. The extracted representations at each stage are then fed into a residual block consisting of two post-normalized $3 \times 3 \times 3$ convolutional layers with instance normalization~\cite{ulyanov2016instance}. The processed features from each stage are then upsampled by using a deconvolutional layer and concatenated with processed features of the preceding stage. The concatenated features are fed into a residual block with aforementioned descriptions. For segmentation, we concatenate the output of the encoder (\textit{i.e.} Swin Transformer) with processed features of the input volume and feed them into a residual block followed by a final $1\times 1\times 1$ convolutional layer with a proper activation function (\textit{i.e.} softmax ) for computing the segmentation probabilities (see Fig.~\ref{fig:fig2} for details of the architecture). 

\section{Pre-training}
We pre-train the Swin UNETR encoder with multiple proxy tasks and formulate it with a multi-objective loss function (Fig.~\ref{fig:fig1}).
The objective of self-supervised representation learning is to encode region of interests (ROI)-aware information of the human body. Inspired by previous works on context reconstruction~\cite{zhou2021models, haghighi2021transferable} and contrastive encoding~\cite{he2020momentum}, we exploit three proxy tasks for medical image representation learning. Three additional projection heads are attached to the encoder during pre-training. Furthermore, the downstream task, e.g. segmentation, fine-tunes the full Swin UNETR model with the projection heads removed. In training, sub-volumes are cropped random regions of the volumetric data. Then, stochastic data augmentations with random rotation and cutout are applied twice to each sub-volume within a mini-batch, resulting in two views of each data.
\subsection{Masked Volume Inpainting}
The cutout augmentation masks out ROIs in the  sub-volume $\mathcal{X}\in\mathbb{R}^{H\times{W}\times{D}\times{C}}$ randomly with volume ratio of $s$. We attach a transpose convolution layer to the encoder as the reconstruction head and denote its output as $\hat{\mathcal{X}}^{\mathcal{M}}$. The reconstruction objective is defined by an $L1$ loss between $\mathcal{X}$ and $\hat{\mathcal{X}}^{\mathcal{M}}$
\begin{equation}
    \mathcal{L}_{inpaint} = \| \mathcal{X} - \hat{\mathcal{X}}^{\mathcal{M}}\|_1,
\end{equation}
The masked volume inpainting is motivated by prior work which focused on 2D images~\cite{pathak2016context}. We extend it to 3D domain to showcase its effectiveness on representation learning of volumetric medical images.

\subsection{Image Rotation}
The rotation prediction task predicts the angle categories by which the input sub-volume is rotated. For simplicity, we employ $R$ classes of $0^{\circ}$, $90^{\circ}$, $180^{\circ}$, $270^{\circ}$ rotations along the $z$-axis. An MLP classification head is used for predicting the softmax probabilities $\hat{y}_{r}$ of rotation categories. Given the ground truth $y_{r}$, a cross-entropy loss is used for rotation prediction task:
\begin{equation}
    \mathcal{L}_{rot} = -\sum^{R}_{r=1} y_r\log(\hat{y}_r),
\label{eq:eq4}
\end{equation}
The 3D rotation and cutout also serves simultaneously as an augmentation transformation for contrastive learning. 

\subsection{Contrastive Coding}
The self-supervised contrastive coding presents promising performance on visual representation learning when transferred to downstream tasks ~\cite{chen2020simple, park2020contrastive}. Given a batch of augmented sub-volumes, the contrastive coding allows for a better representation learning by maximizing the mutual information between positive pairs (augmented samples from same sub-volume), while minimizing that between negative pairs (views from different sub-volumes). The contrastive coding is obtained by attaching a linear layer to the Swin UNETR encoder, which maps each augmented sub-volume to a latent representation $v$. We use cosine similarity as the distance measurement of the encoded representations as defined in ~\cite{chen2020simple}. Formally, the 3D contrastive coding loss between a pair $v_{i}$ and $v_{j}$ is defined as:
\begin{equation}
    \mathcal{L}_{contrast} = 
                -\log\frac{\exp(sim(v_{i}, v_{j})/t)}{\sum^{2N}_{k}1_{k\neq i}\exp(sim(v_{i}, v_{k})/t)},
\label{eq:eq5}
\end{equation}
where $t$ is the measurement of normalized temperature scale. $1$ is the indicator function evaluating to 1 iff $k \neq i$. $sim$ denotes the dot product between normalized embeddings. The contrastive learning loss function strengthens the intra-class compactness as well as the inter-class separability.

\subsection{Loss Function}
Formally, we minimize the total loss function by training Swin UNETR's encoder with multiple pre-training objectives of masked volume inpainting, 3D image rotation \& contrastive coding as follows:
\begin{equation}
    \mathcal{L}_{tot} = \lambda_{1}\mathcal{L}_{inpaint} + \lambda_{2}\mathcal{L}_{contrast} + \lambda_{3}\mathcal{L}_{rot}.
\label{eq:eq6}
\end{equation}
A grid-search hyper-parameter optimization was performed which estimated the optimal values of $\lambda_{1}=\lambda_{2}=\lambda_{3}=1$.

\section{Experiments}
\subsection{Datasets}
\textbf{Pre-training Datasets :} A total of 5 public CT datasets, consisting of 5,050 subjects, are used to construct our pre-training dataset. The corresponding  number of 3D volumes for chest, abdomen and head/neck are 2,018, 1,520 and 1,223 respectively. The collection and source details are presented in the supplementary materials. Existing annotations or labels are \textit{not} utilized from these datasets during the pre-training stage.


\begin{table*}[t]
\scriptsize
\resizebox{\textwidth}{!}{%
\begin{tabular}{l|cccccccccccc|c}
\hline
Methods & \multicolumn{1}{l}{Spl}  
& \multicolumn{1}{l}{RKid} & \multicolumn{1}{l}{LKid} 
& \multicolumn{1}{l}{Gall}  & \multicolumn{1}{l}{Eso} 
& \multicolumn{1}{l}{Liv} & \multicolumn{1}{l}{Sto} 
& \multicolumn{1}{l}{Aor} & \multicolumn{1}{l}{IVC} 
& \multicolumn{1}{l}{Veins}   & \multicolumn{1}{l}{Pan} 
&\multicolumn{1}{l}{AG} & \multicolumn{1}{|l}{Avg.} \\ \hline
SETR NUP~\cite{zheng2020rethinking}
& 0.931 & 0.890                        
& 0.897 & 0.652                        
& 0.760 & 0.952                      
& 0.809 & 0.867   
& 0.745 & 0.717
& 0.719 & 0.620
& 0.796
\\ 
SETR PUP~\cite{zheng2020rethinking}  
& 0.929 & 0.893                        
& 0.892 & 0.649                     
& 0.764 & 0.954                        
& 0.822 & 0.869   
& 0.742 & 0.715
& 0.714 & 0.618
& 0.797
\\ 
SETR MLA~\cite{zheng2021rethinking}    
& 0.930 & 0.889                   
& 0.894 & 0.650                   
& 0.762 & 0.953                      
& 0.819 & 0.872  
& 0.739 & 0.720
& 0.716 & 0.614
& 0.796
\\ 
ASPP~\cite{chen2018encoder}                   
& 0.935 & 0.892                        
& 0.914 & 0.689                        
& 0.760 & 0.953                         
& 0.812 & 0.918   
& 0.807 & 0.695
& 0.720 & 0.629
& 0.811  
\\ 
TransUNet~\cite{chen2021transunet}    
& 0.952 & 0.927                        
& 0.929 & 0.662                        
& 0.757 & 0.969                        
& 0.889 & 0.920   
& 0.833 & 0.791
& 0.775 & 0.637
& 0.838
\\ 
CoTr*~\cite{xie2021cotr}     
& 0.943 & 0.924                        
& 0.929 & 0.687                        
& 0.762 & 0.962                        
& 0.894 & 0.914   
& 0.838 & 0.796
& 0.783 & 0.647
& 0.841
\\ 
CoTr~\cite{xie2021cotr}     
& 0.958 & 0.921                    
& 0.936 & 0.700                        
& 0.764 & 0.963                        
& 0.854 & 0.920  
& 0.838 & 0.787
& 0.775 & 0.694
& 0.844
\\ 
RandomPatch~\cite{tang2021high}     
& 0.963 & 0.912                        
& 0.921 & 0.749                        
& 0.760 & 0.962                        
& 0.870 & 0.889   
& 0.846 & 0.786
& 0.762 & 0.712
& 0.844
\\ 
PaNN~\cite{zhou2019prior}     
& 0.966 & 0.927                       
& 0.952 & 0.732                        
& 0.791 & 0.973                        
& 0.891 & 0.914   
& 0.850 & 0.805
& 0.802 & 0.652
& 0.854
\\
nnUNet~\cite{isensee2021nnu}     
& 0.967 & 0.924                        
& \textbf{0.957} & 0.814                        
& 0.832 & 0.975                        
& 0.925 & 0.928   
& 0.870 & 0.832
& 0.849 & 0.784
& 0.888
\\
UNETR~\cite{hatamizadeh2021unetr}
& 0.972 & 0.942                        
& 0.954 & 0.825                        
& 0.864 & 0.983                        
& 0.945 & 0.948   
& 0.890 & 0.858
& 0.852 & 0.812
& 0.891
\\
\textbf{Swin UNETR}     
& \textbf{0.976} & \textbf{0.958}                        
& 0.956 & \textbf{0.893}                        
& \textbf{0.875} & \textbf{0.985}                        
& \textbf{0.953} & \textbf{0.949}   
& \textbf{0.904} & \textbf{0.899}
& \textbf{0.898} & \textbf{0.846}
& \textbf{0.918}
\\ \hline
\end{tabular}%
}
\\
\caption[caption]%
{Leaderboard\footnotemark Dice results of BTCV challenge on multi-organ segmentation. The proposed method achieves state-of-the-art performance in both free and standard competitions. Note: Spl: spleen, RKid: right kidney, LKid: left kidney, Gall: gallbladder, Eso: esophagus, Liv: liver, Sto: stomach, Aor: aorta, IVC: inferior vena cava, Veins: portal and splenic veins, Pan: pancreas, AG: left and right adrenal glands.}
\label{tab:tab2}
\end{table*}
\footnotetext{\url{https://www.synapse.org/\#!Synapse:syn3193805/wiki/217785/}}

\begin{figure*}[t]
\centering
\includegraphics[width=\textwidth]{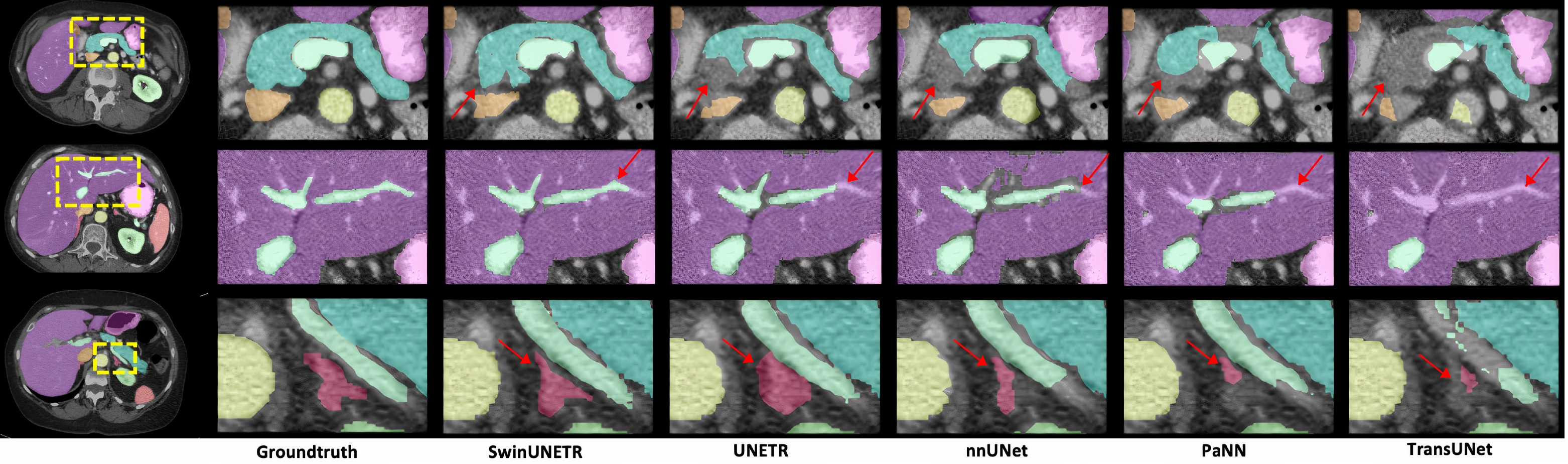}
  \caption{Qualitative visualizations of the proposed Swin UNETR and baseline methods. Three representative subjects are demonstrated. Regions of evident improvements are enlarged to show better details of pancreas (blue), portal vein (light green), and adrenal gland (red).}
  \label{fig:fig4}
\end{figure*}

\textbf{BTCV :} The Beyond the Cranial Vault (BTCV) abdomen challenge dataset ~\cite{landman2015miccai} contains 30 subjects with abdominal CT scans where 13 organs are annotated by interpreters under supervision of radiologists at Vanderbilt University Medical Center. Each CT scan is acquired with contrast enhancement phase at portal venous consists of $80$ to $225$ slices with $512\times 512$ pixels and slice thickness ranging from $1$ to $6$ $mm$. The multi-organ segmentation problem is formulated as a 13 classes segmentation task (see Table~\ref{tab:tab2} for details). The pre-processing pipeline is detailed in supplementary materials. 

\textbf{MSD:} Medical Segmentation Decathlon (MSD) dataset~\cite{antonelli2021medical}
comprises of 10 segmentation tasks from different organs and image modalities. These tasks are designed to feature difficulties across medical images, such as small training sets, unbalanced classes, multi-modality data and small objects. Therefore, the MSD challenge can serve as a comprehensive benchmark to evaluate the generalizability of medical image segmentation methods. The pre-processing pipeline for this dataset is outlined in supplementary materials.

\subsection{Implementation Details}
For pre-training tasks, (1) masked volume inpainting: the ROI dropping rate is set to 30\% (as also used in ~\cite{atito2021sit}); the dropped regions are randomly generated and they sum up to reach overall number of voxels; (2) 3D contrastive coding: a feature size of 512 is used as the embedding size; (3) rotation prediction: the rotation degree is configured to $0^{\circ}$, $90^{\circ}$, $180^{\circ}$, and $270^{\circ}$. We train the model using the AdamW~\cite{loshchilov2018decoupled} optimizer with a warm-up cosine scheduler of 500 iterations. The pre-training experiments use a batch-size of 4 per GPU (with $96\times96\times96$ patch), and initial learning rate of $4e^{-4}$, momentum of 0.9 and decay of $1e^{-5}$ for 450K iterations. Our model is implemented in PyTorch and MONAI\footnote{\href{https://monai.io/}{https://monai.io/}}. A five-fold cross validation strategy is used to train models for all BTCV and MSD experiments. We select the best model in each fold and ensemble their outputs for final segmentation predictions. Detailed training hyperparameters for fine-tuning BTCV and MSD tasks can be found in the supplementary materials. All models are trained on a NVIDIA DGX-1 server.

\subsection{Evaluation Metrics}
The Dice similarity coefficient (Dice) and Hausdorff Distance 95\% (HD95) are used as measurements for experiment results. HD95 calculates 95$^{th}$ percentile of surface distances between ground truth and prediction point sets. Metric formulations are as follows:
\begin{equation}
\textrm{Dice} = \frac{2\sum_{i=1}^{I} Y_{i}\hat{Y}_{i} }{\sum_{i=1}^{I}Y_{i}+ \sum_{i=1}^{I}\hat{Y}_{i}},
\label{eq:dice_score}
\end{equation}
\begin{equation}
\textrm{HD} =\max \{{\max _{y' \in Y'} \min _{\Bar{y}' \in \Bar{Y}'} } \|y'-\Bar{y}'\|, 
\max _{\Bar{y}' \in \Bar{Y}'} \min_{y' \in Y'} \|\Bar{y}'-y'\| \}.
\label{eq:hd_score}
\end{equation}
where $Y$ and $\Bar{Y}$ denote the ground truth and prediction of voxel values. $Y'$ and $\hat{Y}'$ denote ground truth and prediction surface point sets. Surface Dice~\cite{nikolov2018deep} is also used, which is referred as Normalized Surface Distance (NSD) in MSD challenge evaluation. The metric measures the overlap of ground truth and prediction surfaces (with a fixed tolerance) instead of the overlap of two volumes. This provides a measure of agreement between the surfaces of two structures.

\begin{table}[!t]
\centering
\resizebox{.77\linewidth}{!}{
\begin{tabular}{lcccc}
\toprule
\multirow{2}{*}{\textbf{Method}}  &\multirow{3}{*}{\textbf{Rank}}& \multicolumn{2}{c}{\textbf{Average Accuracy}}\\
\cmidrule{3-4}
&  & Dice $\uparrow$ & NSD $\uparrow$  &\\
\midrule
\textbf{Swin UNETR} & \textbf{1} & \textbf{78.68} & \textbf{89.28} 
\\
DiNTS~\cite{he2021dints}   &  2 & 77.93 & 88.68 \\
nnUNet~\cite{isensee2021nnu}   & 3  & 77.89 & 88.09 \\
Models Gen.~\cite{zhou2021models}  & 4  & 76.97 & 87.19 \\
Trans VW~\cite{haghighi2021transferable} & 5 & 76.96 & 87.64& \\
\bottomrule
\end{tabular}}
\caption{Overall performance of top-ranking methods on all 10 segmentation tasks in the MSD public test leaderboard. NSD denotes Normalized Surface Distance.}
\label{tab:alltasks}
\vspace{-1mm}
\end{table}

\begin{table*}[]
 \centering

\resizebox{\textwidth}{!}{
\begin{tabular}{l|rrrr|rrrr|rrr|rrr|rr}
\hline
Organ  & \multicolumn{8}{c|}{Task01 Brain Tumour}  & \multicolumn{6}{c|}{Task03 Liver} & \multicolumn{2}{c}{Task06 Lung} \\
\hline
Metric & Dice1 & Dice2  & Dice3  &Avg.  & NSD1  & NSD2 & NSD3 & Avg. & Dice1 & Dice2 &Avg. & NSD1  & NSD2 &Avg. & Dice1 &NSD1 \\ 
\hline
Kim et al~\cite{kim2019scalable}    & 67.40    & 45.75  & 68.26 & 60.47   & 86.65  & 72.03 & 90.28 & 82.99 & 94.25 & 72.96   & 83.61  & 96.76 & 88.58 & 92.67 & 63.10    & 62.51 \\
Trans VW~\cite{haghighi2021transferable}     & 68.03    & 46.98  & 68.40 & 61.14 & 87.52   & 72.42  & 90.91 & 83.62 & 95.18 & 76.90 & 86.04   & 97.86  & \textbf{92.03} & 94.95 & 74.54    & 76.22 \\
C2FNAS\cite{yu2020c2fnas}       & 67.62    & 48.60  & 69.72 & 61.98   & 87.61  & 72.87 & 91.16 & 83.88 & 94.98 & 72.89   & 83.94  & 98.38 & 89.15 & 93.77  & 70.44    & 72.22 \\
Models Gen.~\cite{zhou2021models}     & 68.03    & 46.98  & 68.40 & 61.14   & 87.52  & 72.42 & 90.91 & 83.62 & 95.72 & \textbf{77.50}   & \textbf{86.61}  & 98.48 & 91.92 & \textbf{95.20} & 74.54    & 76.22\\
nnUNet~\cite{isensee2021nnu}     & 68.04    & 46.81  & 68.46 & 61.10   & 87.51  & 72.47 & 90.78 & 83.59 & \textbf{95.75} & 75.97   & 85.86  & 98.55 & 90.65 & 94.60 & 73.97    & 76.02 \\
DiNTS~\cite{he2021dints}     & 69.28    & 48.65  & 69.75 & 62.56   & \textbf{89.33}  & 73.16 & 91.69 & 84.73 & 95.35 & 74.62   & 84.99  & \textbf{98.69} & 91.02 & 94.86  & 74.75    & 77.02 \\
\hline
Swin UNETR   & \textbf{70.02} &\textbf{52.52} & \textbf{70.51} & \textbf{64.35} & 89.07 & \textbf{80.30} & \textbf{93.46} & \textbf{87.61} & 95.35 & 75.68 & 85.52 & 98.34 & 91.59 & 94.97 & \textbf{76.60} &\textbf{77.40}
\\

\hline

\end{tabular}%
}
\label{tab:task01}
\end{table*}

\begin{table*}[]
 \centering
\resizebox{\textwidth}{!}{
\begin{tabular}{l|rrr|rrr|rrr|rrr|rr|rr}
\hline
Organ  & \multicolumn{6}{c|}{Task07 Pancreas} & \multicolumn{6}{c|}{Task08 Hepatic Vessel} & \multicolumn{2}{c|}{Task09 Spleen} & \multicolumn{2}{c}{Task10 Colon} \\
\hline
Metric  & Dice1  & Dice2  & Avg.  & NSD1 & NSD2 & Avg.  & Dice1  & Dice2  & Avg.  & NSD1 & NSD2 & Avg. & Dice1 & NSD1 & Dice1 & NSD1\\ 
\hline
Kim et al~\cite{kim2019scalable}    & 80.61    & 51.75  & 66.18 & 95.83   & 73.09  & 84.46 & 62.34 & 68.63   & 65.49  & 83.22 & 78.43 & 80.83 & 91.92    & 94.83 & 49.32    & 62.21\\
Trans VW~\cite{haghighi2021transferable}     & 81.42    & 51.08  & 66.25 & 96.07 & 70.13   & 83.10 & 65.80 & 71.44 & 68.62   & 84.01  & 80.15 & 82.08 & 97.35    & 99.87 & 51.47    & 60.53\\
C2FNAS\cite{yu2020c2fnas}       & 80.76    & 54.41  & 67.59 & 96.16   & 75.58  & 85.87 & 64.30 & 71.00   & 67.65  & 83.78 & 80.66 & 82.22  & 96.28    & 97.66 & 58.90    & \textbf{72.56}\\
Models Gen.~\cite{zhou2021models}     & 81.36    & 50.36  & 65.86 & 96.16   & 70.02  & 83.09 & 65.80 & 71.44   & 68.62  & 84.01 & 80.15 & 82.08  & 97.35    & 99.87 & 51.47    & 60.53\\
nnUNet~\cite{isensee2021nnu}     & 81.64    & 52.78  & 67.21 & 96.14   & 71.47  & 83.81 & \textbf{66.46} & 71.78   & \textbf{69.12}  & 84.43 & 80.72 & 82.58 & \textbf{97.43}    & \textbf{99.89} & 58.33    & 68.43\\
DiNTS~\cite{he2021dints}     & 81.02    & 55.35  & 68.19 & 96.26   & 75.90  & 86.08 & 64.50 & 71.76   & 68.13  & 83.98 & 81.03 & 82.51  & 96.98    & 99.83 & 59.21    & 70.34\\
\hline
Swin UNETR   & \textbf{81.85} &\textbf{58.21} & \textbf{70.71} & \textbf{96.57} &\textbf{79.10} & \textbf{87.84} & 65.69 & \textbf{72.20} &68.95 & \textbf{84.83} & \textbf{81.62} & \textbf{83.23}  & 96.99 & 99.84 & \textbf{59.45} & 70.89
\\

\hline

\end{tabular}%
}
\caption{MSD test dataset performance comparison of Dice and NSD. Benchmarks obtained from MSD test leaderboard\protect\footnotemark.}
\label{tab:task07}
\end{table*}
\footnotetext{\url{https://decathlon-10.grand-challenge.org/evaluation/challenge/leaderboard/}}

\subsection{Results}
\subsubsection{BTCV Multi-organ Segmentation Challenge}
We extensively compare the benchmarks of our model with baselines. The published leaderboard evaluation is shown in Table~\ref{tab:tab2}. Compared with other top submissions, the proposed Swin UNETR achieves the best performance. We obtain the state-of-the-art Dice of 0.908, outperforming the second, third and fourth top-ranked baselines by 1.6\%, 2.0\% and 2.4\% on average of 13 organs, respectively. Distinct improvements can be specifically observed for organs that are smaller in size, such as splenic and portal veins of 3.6\%  against prior state-of-the-art method, pancreas of 1.6\%, and adrenal glands of 3.8\%. Moderate improvements are observed in other organs. The representative samples in Fig.~\ref{fig:fig4} demonstrate the success of identifying organ details by Swin UNETR. Our method detects the pancreas tail (row 1), and branches in the portal vein (row 2) in  Fig.~\ref{fig:fig4}, where other methods under segment parts of each tissue. In addition, our method demonstrates distinct improvement in segmentation of adrenal glands (row 3).

\subsubsection{Segmentation Results on MSD}
\begin{figure}[t]
\includegraphics[width=\linewidth]{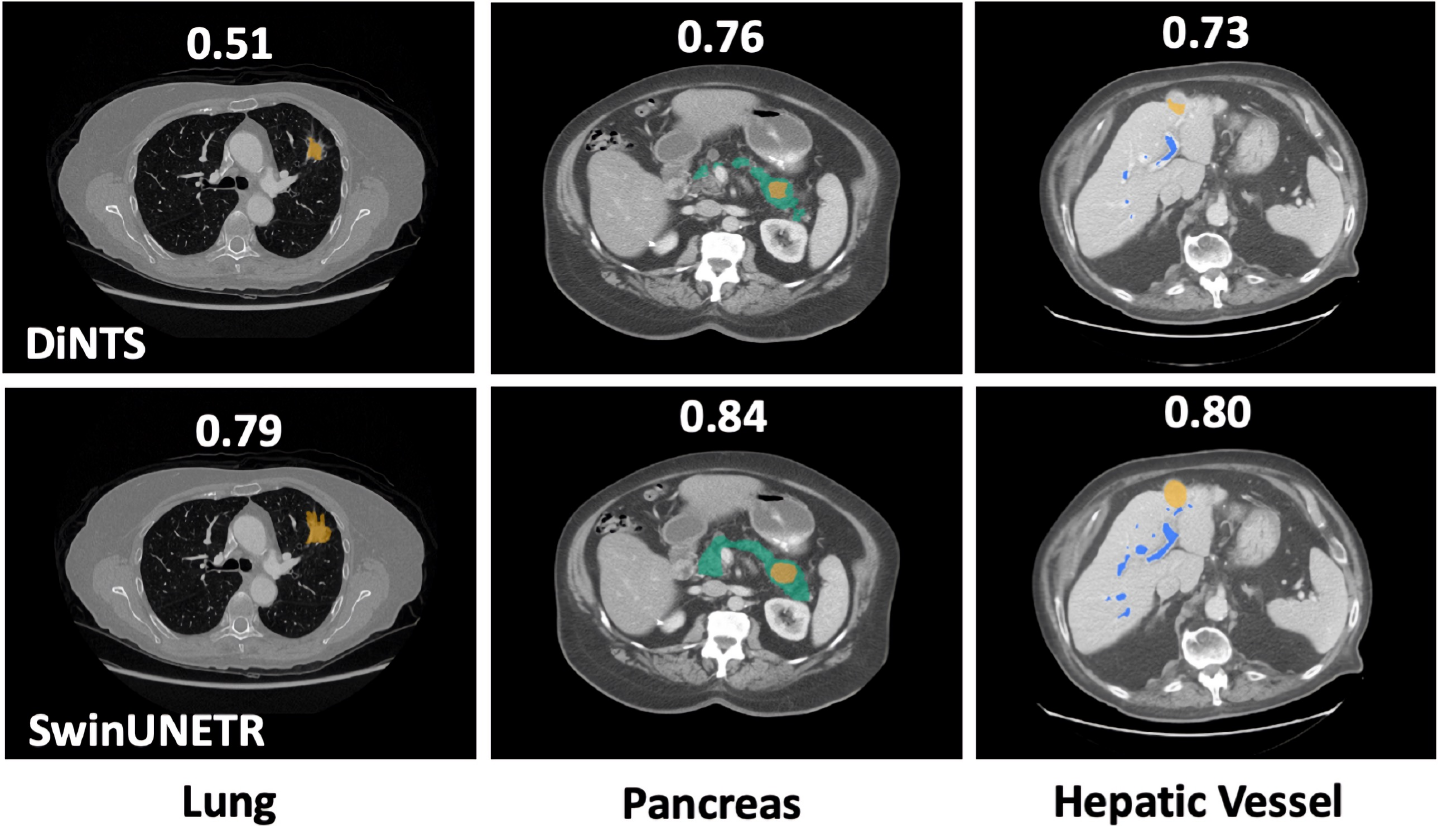}
  \caption{Qualitative results of representative MSD CT tasks. Average Dice values are illustrated on top of each image. Our model demonstrates more accurate performance in comparison to DiNTS for both organ and tumor segmentation across different tasks.}
  \label{fig:fig5}
\end{figure}

 The overall MSD results per task and ranking from the challenge leaderboard are shown in Table.~\ref{tab:alltasks}. The proposed Swin UNETR achieves state-of-the-art performance in Task01 BrainTumour, Task06 Lung, Task07 Pancreas, and Task10 Colon. The results are comparable for Task02 Heart, Task03 Liver, Task04 Hippocampus, Task05 Prostate, Task08 HepaticVessel and Task09 Spleen. Overall, Swin UNETR presents the best average Dice of 78.68\% across all ten tasks and achieves the top ranking in the MSD leaderboard. The detail number of multiple tasks are shown in Table~\ref{tab:task07}. Qualitative visualization can be observed in Fig.~\ref{fig:fig5}. Swin UNETR with self-supervised pre-training demonstrates visually better segmentation results in the CT tasks. The pre-trained weights are only used for fine-tuning CT tasks including Liver, Lung, Pancreas, HepaticVessel, Spleen, and Colon. For MRI tasks: Brain Tumour, Heart, Hippocampus, Prostate, experiments are trained from scratch because of the domain gap between CT and MRI images. Due to space limitations, we present the MSD test benchmarks for the remaining three MRI tasks in the supplementary materials.
 
\begin{table}[!t]
\centering
\resizebox{.71\linewidth}{!}{
\begin{tabular}{lccc}
\toprule
\multirow{2}{*}{\textbf{Loss Function}}  &  \multicolumn{3}{c}{\textbf{Average Accuracy}}\\
\cmidrule{2-4}
&  Dice $\uparrow$ & HD $\downarrow$ \\
\midrule
 \ \ \ \ \ Scratch & $83.43$ & $42.36$   \\
\midrule
        \ \ $\mathcal{L_\text{rot}}$ & $83.56$ & $36.19$ \\
    \ \ $\mathcal{L_\text{contrast}}$ & $83.67$ & $38.81$ \\
            \ \ $\mathcal{L_\text{inpaint}}$ & $83.85$ & $28.94$ \\
                        \ \ $\mathcal{L_\text{inpaint}} + \mathcal{L_\text{rot}}$ & $84.01$ & $26.06$ \\
                        \ \ $\mathcal{L_\text{inpaint}} + \mathcal{L_\text{contrast}}$ & $84.45$ & $24.37$ \\
                        \ \ $\mathcal{L_\text{inpaint}} + \mathcal{L_\text{contrast}+} \mathcal{L_\text{rot}}$ & $\mathbf{84.72}$ & $\mathbf{20.03}$ \\
\bottomrule
\end{tabular}}
\vspace{1mm}

\caption{Ablation study of the effectiveness of each objective function in the proposed pre-training loss. HD denotes Hausdorff Distance. Experiments on fine-tuning the BTCV dataset.
}
\vspace{0.2mm}
\label{table:loss_ablation_table}
\end{table}

\subsection{Ablation Study}
\subsubsection{Efficacy of Pre-training}
\begin{figure}[t]
\includegraphics[width=\linewidth]{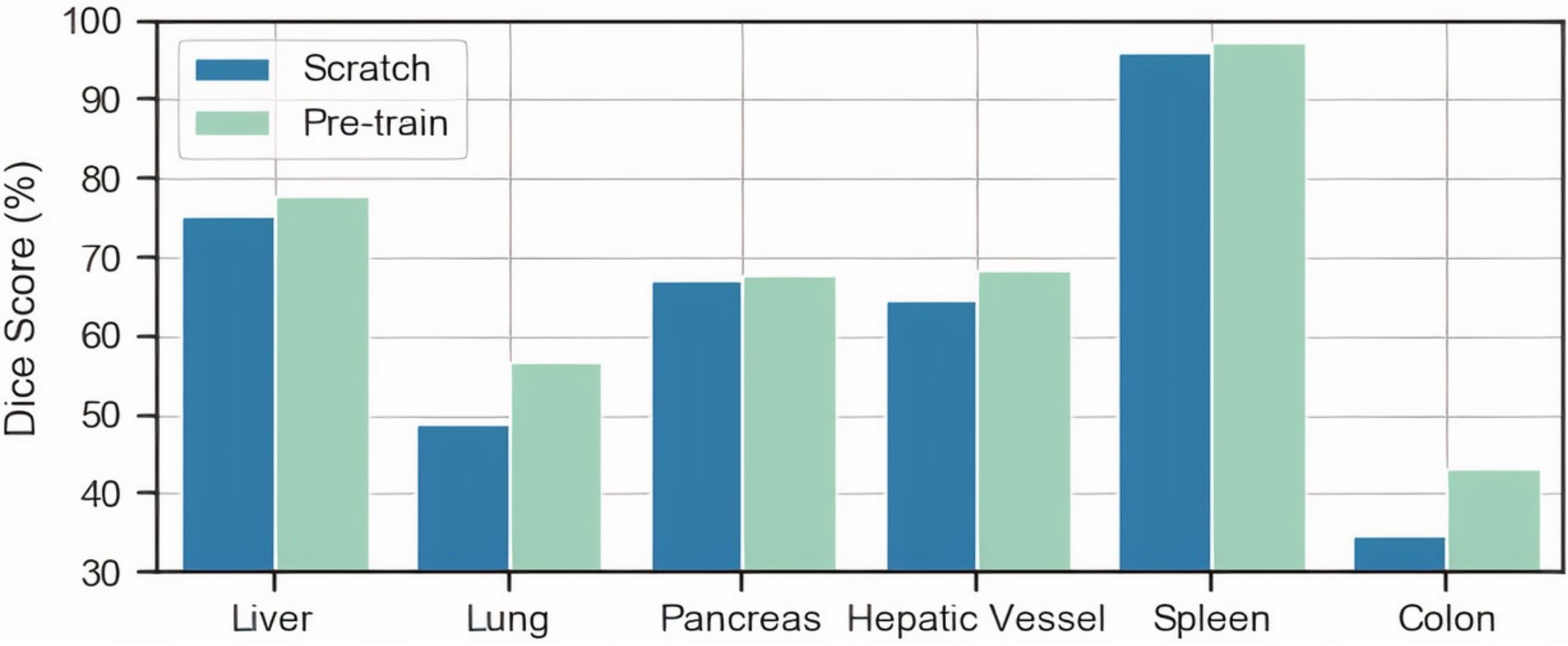}
  \caption{The indication of Dice gap between using pre-training (Green) and scratch model (Blue) on MSD CT tasks validation set.}
  \label{fig:fig6}
\end{figure}

A comparison of all MSD CT tasks using pre-trained model against training from scratch can be observed in Fig.~\ref{fig:fig6}. Distinct improvement can be observed for Task03 Liver, Dice of 77.77\% comparing to 75.27\%. Task08 Hepatic Vessel achieves 68.52\% against 64.63\%. Task10 Colon shows the largest improvement, from 34.83\% to 43.38\%. Task07 Pancreas and Task09 Spleen both achieve significant improvement from 67.12\% to 67.82\%, and 96.05\% to 97.32\% respectively. 

\subsubsection{Reduce Manual Labeling Efforts}
\begin{figure}[t]
\includegraphics[width=\linewidth]{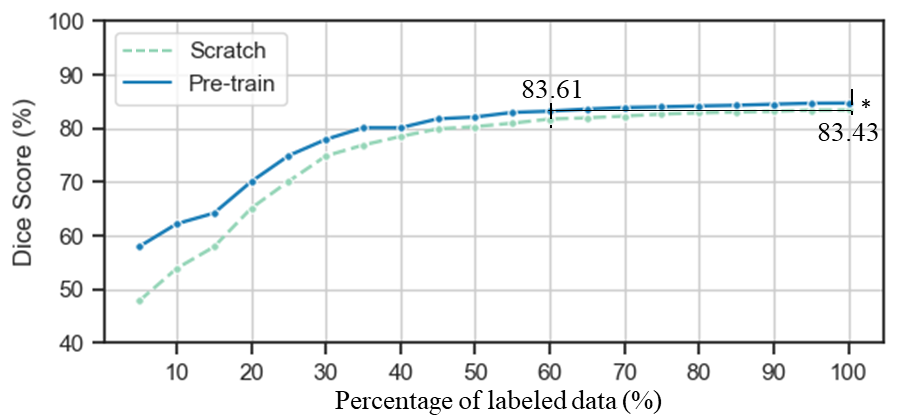}
  \caption{Data-efficient performance on BTCV test dataset. Significance under Wilcoxon Signed Rank test, $*: p < 0.001$.}
  \label{fig:fig7}
\end{figure}

Fig.~\ref{fig:fig7} demonstrates the comparison results of fine-tuning using a subset of BTCV dataset. We show using 10\% of labeled data, experiments with pre-training weights achieve approximately 10\% improvement comparing to training from scratch. On employing all labeled data, the self-supervised pre-training shows 1.3\% higher average Dice. The Dice number 83.13 of learning from scratch with entire dataset can be achieved by using pre-trained Swin UNETR with 60\% data. Fig.~\ref{fig:fig7} indicates that our approach can reduce the annotation effort by at least 40\% for BTCV task. 
\subsubsection{Size of Pre-training Dataset}
\begin{figure}[t]
\includegraphics[width=\linewidth]{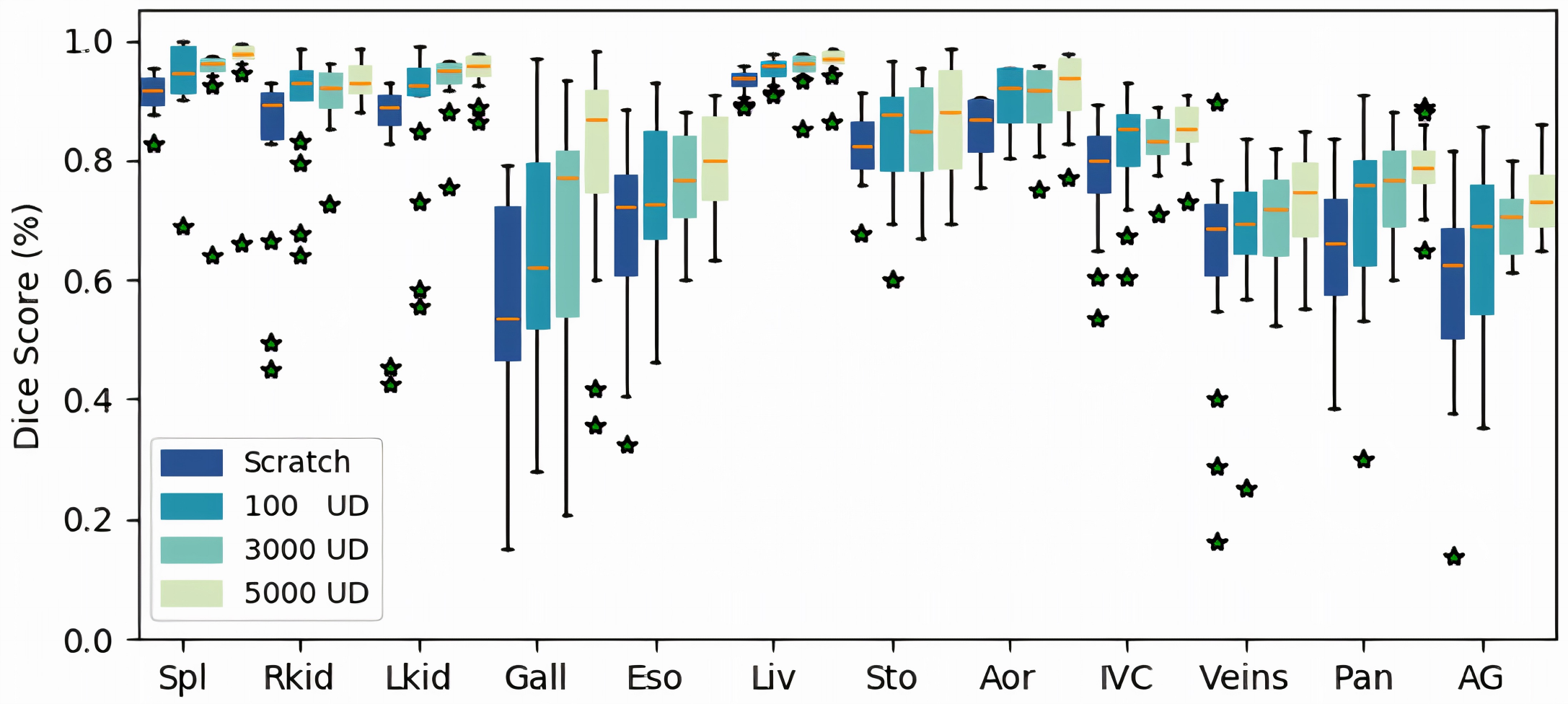}
  \caption{Pre-trained weights using 100, 3000 and 5000 scans are compared for fine-tuning on the BTCV dataset for each organ. }
  \label{fig:fig8}
\end{figure}

We perform organ-wise study on BTCV dataset by using pre-trained weights of smaller unlabeled data. In  Fig.~\ref{fig:fig8}, the fine-tuning results are obtained from pre-training 100, 3,000, and 5,000 scans. We observe that Swin UNETR is robust with respect to the total number of CT scans trained.  Fig.~\ref{fig:fig8} demonstrates the proposed model can benefit from larger pre-training datasets with increasing size of unlabeled data.

\subsubsection{Efficacy of Self-Supervised Objectives}
We perform empirical study on pre-training with different combinations of self-supervised objectives. As shown in Table~\ref{table:loss_ablation_table}, on BTCV test set, using pre-trained weights by inpainting achieves the highest improvement at single task modeling. On pairing tasks, inpainting and contrastive learning show Dice of 84.45\% and Hausdorff Distance (HD) of 24.37. Overall, employing all proxy tasks achieves best Dice of 84.72\%.

\section{Discussion and Limitations}
\vspace*{-0.8mm}
Our state-of-the-art results on the test leaderboards of MSD and BTCV datasets validate the effectiveness of the proposed self-supervised learning framework in taking the advantage of large number of available medical images without the need of annotation effort. Subsequently, fine-tuning the pretrained Swin UNETR model achieves higher accuracy, improves the convergence speed, and reduces the annotation effort in comparison to training with randomly initialized weights from scratch. 
Our framework is scalable and can be easily extended with more proxy tasks and augmentation transformations. Meanwhile, the pre-trained encoder can benefit the transfer learning of various medical imaging analysis tasks, such as classification and detection. In MSD pancreas segmentation task, Swin UNETR with pre-trained weights outperforms AutoML algorithms such as DiNTS~\cite{he2021dints} and C2FNAS~\cite{yu2020c2fnas} that are specifically designed for searching the optimal network architectures on the same segmentation task. Currently, Swin UNETR has only been pre-trained using CT images, and our experiments have not demonstrated enough transferability when applied directly to other medical imaging modalities such as MRI. This is mainly due to obvious domain gaps and different number of input channels that are specific to each modality. As a result, this is a potential direction that should be studied in future efforts.
\vspace*{-1.7mm}
\section{Conclusions}
In this work, we present a novel framework for self-supervised pre-training of 3D medical images. Inspired by merging feature maps at scales, we built the Swin UNETR by exploiting transformer-encoded spatial representations into convolution-based decoders. By proposing the first transformer-based 3D medical image pre-training, we leverage the power of Swin Transformer encoder for fine-tuning segmentation tasks. Swin UNETR with self-supervised pre-training achieves the state-of-the-art performance on the BTCV multi-organ segmentation challenge and MSD challenge. Particularly, we presented the large-scale CT pre-training with 5,050 volumes, by combining multiple publicly available datasets and diversities of anatomical ROIs.

{\small
\bibliographystyle{ieee_fullname}
\bibliography{egbib}

\begin{thebibliography}{10}\itemsep=-1pt

\bibitem{antonelli2021medical}
Michela Antonelli, Annika Reinke, Spyridon Bakas, Keyvan Farahani, Bennett~A
  Landman, Geert Litjens, Bjoern Menze, Olaf Ronneberger, Ronald~M Summers,
  Bram van Ginneken, et~al.
\newblock The medical segmentation decathlon.
\newblock {\em arXiv preprint arXiv:2106.05735}, 2021.

\bibitem{armato2011lung}
Samuel~G Armato~III, Geoffrey McLennan, Luc Bidaut, Michael~F McNitt-Gray,
  Charles~R Meyer, Anthony~P Reeves, Binsheng Zhao, Denise~R Aberle, Claudia~I
  Henschke, Eric~A Hoffman, et~al.
\newblock The lung image database consortium (lidc) and image database resource
  initiative (idri): a completed reference database of lung nodules on ct
  scans.
\newblock {\em Medical physics}, 38(2):915--931, 2011.

\bibitem{atito2021sit}
Sara Atito, Muhammad Awais, and Josef Kittler.
\newblock Sit: Self-supervised vision transformer.
\newblock {\em arXiv preprint arXiv:2104.03602}, 2021.

\bibitem{azizi2021big}
Shekoofeh Azizi, Basil Mustafa, Fiona Ryan, Zachary Beaver, Jan Freyberg,
  Jonathan Deaton, Aaron Loh, Alan Karthikesalingam, Simon Kornblith, Ting
  Chen, et~al.
\newblock Big self-supervised models advance medical image classification.
\newblock {\em Proceedings of the IEEE/CVF International Conference on Computer
  Vision}, 2021.

\bibitem{cao2021swin}
Hu Cao, Yueyue Wang, Joy Chen, Dongsheng Jiang, Xiaopeng Zhang, Qi Tian, and
  Manning Wang.
\newblock Swin-unet: Unet-like pure transformer for medical image segmentation.
\newblock {\em arXiv preprint arXiv:2105.05537}, 2021.

\bibitem{caron2021emerging}
Mathilde Caron, Hugo Touvron, Ishan Misra, Herv{\'e} J{\'e}gou, Julien Mairal,
  Piotr Bojanowski, and Armand Joulin.
\newblock Emerging properties in self-supervised vision transformers.
\newblock {\em Proceedings of the IEEE/CVF International Conference on Computer
  Vision}, 2021.

\bibitem{chen2021transunet}
Jieneng Chen, Yongyi Lu, Qihang Yu, Xiangde Luo, Ehsan Adeli, Yan Wang, Le Lu,
  Alan~L Yuille, and Yuyin Zhou.
\newblock Transunet: Transformers make strong encoders for medical image
  segmentation.
\newblock {\em arXiv preprint arXiv:2102.04306}, 2021.

\bibitem{chen2019self}
Liang Chen, Paul Bentley, Kensaku Mori, Kazunari Misawa, Michitaka Fujiwara,
  and Daniel Rueckert.
\newblock Self-supervised learning for medical image analysis using image
  context restoration.
\newblock {\em Medical image analysis}, 58:101539, 2019.

\bibitem{chen2018encoder}
Liang-Chieh Chen, Yukun Zhu, George Papandreou, Florian Schroff, and Hartwig
  Adam.
\newblock Encoder-decoder with atrous separable convolution for semantic image
  segmentation.
\newblock In {\em European Conference on Computer Vision}, 2018.

\bibitem{deeplabv3plus2018}
Liang-Chieh Chen, Yukun Zhu, George Papandreou, Florian Schroff, and Hartwig
  Adam.
\newblock Encoder-decoder with atrous separable convolution for semantic image
  segmentation.
\newblock {\em arXiv:1802.02611}, 2018.

\bibitem{chen2019med3d}
Sihong Chen, Kai Ma, and Yefeng Zheng.
\newblock Med3d: Transfer learning for 3d medical image analysis.
\newblock {\em arXiv preprint arXiv:1904.00625}, 2019.

\bibitem{chen2020simple}
Ting Chen, Simon Kornblith, Mohammad Norouzi, and Geoffrey Hinton.
\newblock A simple framework for contrastive learning of visual
  representations.
\newblock In {\em International conference on machine learning}, 2020.

\bibitem{chen2021empirical}
Xinlei Chen, Saining Xie, and Kaiming He.
\newblock An empirical study of training self-supervised vision transformers.
\newblock {\em Proceedings of the IEEE/CVF International Conference on Computer
  Vision}, 2021.

\bibitem{cheng2021per}
Bowen Cheng, Alexander~G Schwing, and Alexander Kirillov.
\newblock Per-pixel classification is not all you need for semantic
  segmentation.
\newblock {\em arXiv preprint arXiv:2107.06278}, 2021.

\bibitem{cciccek20163d}
{\"O}zg{\"u}n {\c{C}}i{\c{c}}ek, Ahmed Abdulkadir, Soeren~S Lienkamp, Thomas
  Brox, and Olaf Ronneberger.
\newblock 3d u-net: learning dense volumetric segmentation from sparse
  annotation.
\newblock In {\em International conference on medical image computing and
  computer-assisted intervention}, 2016.

\bibitem{dai2021up}
Zhigang Dai, Bolun Cai, Yugeng Lin, and Junying Chen.
\newblock Up-detr: Unsupervised pre-training for object detection with
  transformers.
\newblock In {\em Proceedings of the IEEE/CVF Conference on Computer Vision and
  Pattern Recognition}, 2021.

\bibitem{deng2009imagenet}
Jia Deng, Wei Dong, Richard Socher, Li-Jia Li, Kai Li, and Li Fei-Fei.
\newblock Imagenet: A large-scale hierarchical image database.
\newblock In {\em Proceedings of the IEEE/CVF Conference on Computer Vision and
  Pattern Recognition}, 2009.

\bibitem{desai2020chest}
Shivang Desai, Ahmad Baghal, Thidathip Wongsurawat, Piroon Jenjaroenpun, Thomas
  Powell, Shaymaa Al-Shukri, Kim Gates, Phillip Farmer, Michael Rutherford,
  Geri Blake, et~al.
\newblock Chest imaging representing a covid-19 positive rural us population.
\newblock {\em Scientific data}, 7(1):1--6, 2020.

\bibitem{devlin2018bert}
Jacob Devlin, Ming-Wei Chang, Kenton Lee, and Kristina Toutanova.
\newblock Bert: Pre-training of deep bidirectional transformers for language
  understanding.
\newblock {\em arXiv preprint arXiv:1810.04805}, 2018.

\bibitem{dosovitskiy2020image}
Alexey Dosovitskiy, Lucas Beyer, Alexander Kolesnikov, Dirk Weissenborn,
  Xiaohua Zhai, Thomas Unterthiner, Mostafa Dehghani, Matthias Minderer, Georg
  Heigold, Sylvain Gelly, et~al.
\newblock An image is worth 16x16 words: Transformers for image recognition at
  scale.
\newblock In {\em International Conference on Learning Representations}, 2020.

\bibitem{gidaris2018unsupervised}
Spyros Gidaris, Praveer Singh, and Nikos Komodakis.
\newblock Unsupervised representation learning by predicting image rotations.
\newblock In {\em International Conference on Learning Representations}, 2018.

\bibitem{grossberg2018imaging}
Aaron~J Grossberg, Abdallah~SR Mohamed, Hesham Elhalawani, William~C Bennett,
  Kirk~E Smith, Tracy~S Nolan, Bowman Williams, Sasikarn Chamchod, Jolien
  Heukelom, Michael~E Kantor, et~al.
\newblock Imaging and clinical data archive for head and neck squamous cell
  carcinoma patients treated with radiotherapy.
\newblock {\em Scientific data}, 5(1):1--10, 2018.

\bibitem{haghighi2021transferable}
Fatemeh Haghighi, Mohammad Reza~Hosseinzadeh Taher, Zongwei Zhou, Michael~B
  Gotway, and Jianming Liang.
\newblock Transferable visual words: Exploiting the semantics of anatomical
  patterns for self-supervised learning.
\newblock {\em IEEE transactions on medical imaging}, 2021.

\bibitem{hatamizadeh2021unetr}
Ali Hatamizadeh, Yucheng Tang, Vishwesh Nath, Dong Yang, Andriy Myronenko,
  Bennett Landman, Holger~R Roth, and Daguang Xu.
\newblock Unetr: Transformers for 3d medical image segmentation.
\newblock In {\em Proceedings of the IEEE/CVF Winter Conference on Applications
  of Computer Vision}, pages 574--584, 2022.

\bibitem{he2020momentum}
Kaiming He, Haoqi Fan, Yuxin Wu, Saining Xie, and Ross Girshick.
\newblock Momentum contrast for unsupervised visual representation learning.
\newblock In {\em Proceedings of the IEEE/CVF Conference on Computer Vision and
  Pattern Recognition}, 2020.

\bibitem{he2016deep}
Kaiming He, Xiangyu Zhang, Shaoqing Ren, and Jian Sun.
\newblock Deep residual learning for image recognition.
\newblock In {\em Proceedings of the IEEE/CVF Conference on Computer Vision and
  Pattern Recognition}, 2016.

\bibitem{he2021dints}
Yufan He, Dong Yang, Holger Roth, Can Zhao, and Daguang Xu.
\newblock Dints: Differentiable neural network topology search for 3d medical
  image segmentation.
\newblock In {\em Proceedings of the IEEE/CVF Conference on Computer Vision and
  Pattern Recognition}, 2021.

\bibitem{isensee2021nnu}
Fabian Isensee, Paul~F Jaeger, Simon~AA Kohl, Jens Petersen, and Klaus~H
  Maier-Hein.
\newblock nnu-net: a self-configuring method for deep learning-based biomedical
  image segmentation.
\newblock {\em Nature Methods}, 18(2):203--211, 2021.

\bibitem{johnson2008accuracy}
C~Daniel Johnson, Mei-Hsiu Chen, Alicia~Y Toledano, Jay~P Heiken, Abraham
  Dachman, Mark~D Kuo, Christine~O Menias, Betina Siewert, Jugesh~I Cheema,
  Richard~G Obregon, et~al.
\newblock Accuracy of ct colonography for detection of large adenomas and
  cancers.
\newblock {\em New England Journal of Medicine}, 359(12):1207--1217, 2008.

\bibitem{jose2021medical}
JM Jose and P Oza.
\newblock Medical transformer: gated axial-attention for medical image
  segmentation.
\newblock In {\em International Conference on Medical Image Computing and
  Computer Assisted Intervention}, 2021.

\bibitem{kim2019scalable}
Sungwoong Kim, Ildoo Kim, Sungbin Lim, Woonhyuk Baek, Chiheon Kim, Hyungjoo
  Cho, Boogeon Yoon, and Taesup Kim.
\newblock Scalable neural architecture search for 3d medical image
  segmentation.
\newblock In {\em International Conference on Medical Image Computing and
  Computer-Assisted Intervention}, 2019.

\bibitem{landman2015miccai}
B Landman, Z Xu, J Igelsias, M Styner, T Langerak, and A Klein.
\newblock Miccai multi-atlas labeling beyond the cranial vault--workshop and
  challenge.
\newblock In {\em Proc. MICCAI Multi-Atlas Labeling Beyond Cranial
  Vault—Workshop Challenge}, 2015.

\bibitem{liang2021swinir}
Jingyun Liang, Jiezhang Cao, Guolei Sun, Kai Zhang, Luc Van~Gool, and Radu
  Timofte.
\newblock Swinir: Image restoration using swin transformer.
\newblock In {\em Proceedings of the IEEE/CVF International Conference on
  Computer Vision}, 2021.

\bibitem{lin2021ds}
Ailiang Lin, Bingzhi Chen, Jiayu Xu, Zheng Zhang, and Guangming Lu.
\newblock Ds-transunet: Dual swin transformer u-net for medical image
  segmentation.
\newblock {\em arXiv preprint arXiv:2106.06716}, 2021.

\bibitem{lin2017feature}
Tsung-Yi Lin, Piotr Doll{\'a}r, Ross Girshick, Kaiming He, Bharath Hariharan,
  and Serge Belongie.
\newblock Feature pyramid networks for object detection.
\newblock In {\em In Proceedings of the IEEE/CVF Interna-tional Conference on
  Computer Vision}, 2017.

\bibitem{liu2021swin}
Ze Liu, Yutong Lin, Yue Cao, Han Hu, Yixuan Wei, Zheng Zhang, Stephen Lin, and
  Baining Guo.
\newblock Swin transformer: Hierarchical vision transformer using shifted
  windows.
\newblock {\em Proceedings of the IEEE/CVF International Conference on Computer
  Vision}, 2021.

\bibitem{liu2021video}
Ze Liu, Jia Ning, Yue Cao, Yixuan Wei, Zheng Zhang, Stephen Lin, and Han Hu.
\newblock Video swin transformer.
\newblock {\em arXiv preprint arXiv:2106.13230}, 2021.

\bibitem{loshchilov2018decoupled}
Ilya Loshchilov and Frank Hutter.
\newblock Decoupled weight decay regularization.
\newblock In {\em International Conference on Learning Representations}, 2018.

\bibitem{nikolov2018deep}
Stanislav Nikolov, Sam Blackwell, Alexei Zverovitch, Ruheena Mendes, Michelle
  Livne, Jeffrey De~Fauw, Yojan Patel, Clemens Meyer, Harry Askham, Bernardino
  Romera-Paredes, et~al.
\newblock Deep learning to achieve clinically applicable segmentation of head
  and neck anatomy for radiotherapy.
\newblock {\em arXiv preprint arXiv:1809.04430}, 2018.

\bibitem{noroozi2016unsupervised}
Mehdi Noroozi and Paolo Favaro.
\newblock Unsupervised learning of visual representations by solving jigsaw
  puzzles.
\newblock In {\em European conference on computer vision}, 2016.

\bibitem{oord2018representation}
Aaron van~den Oord, Yazhe Li, and Oriol Vinyals.
\newblock Representation learning with contrastive predictive coding.
\newblock {\em arXiv preprint arXiv:1807.03748}, 2018.

\bibitem{park2020contrastive}
Taesung Park, Alexei~A Efros, Richard Zhang, and Jun-Yan Zhu.
\newblock Contrastive learning for unpaired image-to-image translation.
\newblock In {\em European Conference on Computer Vision}, 2020.

\bibitem{pathak2016context}
Deepak Pathak, Philipp Krahenbuhl, Jeff Donahue, Trevor Darrell, and Alexei~A
  Efros.
\newblock Context encoders: Feature learning by inpainting.
\newblock In {\em Proceedings of the IEEE/CVF Conference on Computer Vision and
  Pattern Recognition}, 2016.

\bibitem{raghu2021vision}
Maithra Raghu, Thomas Unterthiner, Simon Kornblith, Chiyuan Zhang, and Alexey
  Dosovitskiy.
\newblock Do vision transformers see like convolutional neural networks?
\newblock {\em arXiv preprint arXiv:2108.08810}, 2021.

\bibitem{raghu2019transfusion}
Maithra Raghu, Chiyuan Zhang, Jon Kleinberg, and Samy Bengio.
\newblock Transfusion: Understanding transfer learning for medical imaging.
\newblock {\em Advances in Neural Information Processing Systems}, 2019.

\bibitem{roth2018multi}
Holger~R Roth, Chen Shen, Hirohisa Oda, Takaaki Sugino, Masahiro Oda, Yuichiro
  Hayashi, Kazunari Misawa, and Kensaku Mori.
\newblock A multi-scale pyramid of 3d fully convolutional networks for
  abdominal multi-organ segmentation.
\newblock In {\em International conference on medical image computing and
  computer-assisted intervention}, 2018.

\bibitem{setio2017validation}
Arnaud Arindra~Adiyoso Setio, Alberto Traverso, Thomas De~Bel, Moira~SN Berens,
  Cas Van Den~Bogaard, Piergiorgio Cerello, Hao Chen, Qi Dou, Maria~Evelina
  Fantacci, Bram Geurts, et~al.
\newblock Validation, comparison, and combination of algorithms for automatic
  detection of pulmonary nodules in computed tomography images: the luna16
  challenge.
\newblock {\em Medical image analysis}, 42:1--13, 2017.

\bibitem{simpson2019large}
Amber~L Simpson, Michela Antonelli, Spyridon Bakas, Michel Bilello, Keyvan
  Farahani, Bram Van~Ginneken, Annette Kopp-Schneider, Bennett~A Landman, Geert
  Litjens, Bjoern Menze, et~al.
\newblock A large annotated medical image dataset for the development and
  evaluation of segmentation algorithms.
\newblock {\em arXiv preprint arXiv:1902.09063}, 2019.

\bibitem{singh2018sniper}
Bharat Singh, Mahyar Najibi, and Larry~S Davis.
\newblock Sniper: Efficient multi-scale training.
\newblock {\em Advances in Neural Information Processing Systems},
  31:9310--9320, 2018.

\bibitem{NEURIPS2020_d2dc6368}
Aiham Taleb, Winfried Loetzsch, Noel Danz, Julius Severin, Thomas Gaertner,
  Benjamin Bergner, and Christoph Lippert.
\newblock 3d self-supervised methods for medical imaging.
\newblock In {\em Advances in Neural Information Processing Systems}, 2020.

\bibitem{tang2021body}
Yucheng Tang, Riqiang Gao, Shizhong Han, Yunqiang Chen, Dashan Gao, Vishwesh
  Nath, Camilo Bermudez, Michael~R Savona, Shunxing Bao, Ilwoo Lyu, et~al.
\newblock Body part regression with self-supervision.
\newblock {\em IEEE Transactions on Medical Imaging}, 40(5):1499--1507, 2021.

\bibitem{tang2021high}
Yucheng Tang, Riqiang Gao, Ho~Hin Lee, Shizhong Han, Yunqiang Chen, Dashan Gao,
  Vishwesh Nath, Camilo Bermudez, Michael~R Savona, Richard~G Abramson, et~al.
\newblock High-resolution 3d abdominal segmentation with random patch network
  fusion.
\newblock {\em Medical Image Analysis}, 69:101894, 2021.

\bibitem{ulyanov2016instance}
Dmitry Ulyanov, Andrea Vedaldi, and Victor Lempitsky.
\newblock Instance normalization: The missing ingredient for fast stylization.
\newblock {\em arXiv preprint arXiv:1607.08022}, 2016.

\bibitem{wang2021self}
Xiaosong Wang, Ziyue Xu, Leo Tam, Dong Yang, and Daguang Xu.
\newblock Self-supervised image-text pre-training with mixed data in chest
  x-rays.
\newblock {\em arXiv preprint arXiv:2103.16022}, 2021.

\bibitem{xie2021cotr}
Yutong Xie, Jianpeng Zhang, Chunhua Shen, and Yong Xia.
\newblock Cotr: Efficiently bridging cnn and transformer for 3d medical image
  segmentation.
\newblock {\em International conference on medical image computing and
  computer-assisted intervention}, 2021.

\bibitem{xie2021self}
Zhenda Xie, Yutong Lin, Zhuliang Yao, Zheng Zhang, Qi Dai, Yue Cao, and Han Hu.
\newblock Self-supervised learning with swin transformers.
\newblock {\em arXiv preprint arXiv:2105.04553}, 2021.

\bibitem{xu2021levit}
Guoping Xu, Xingrong Wu, Xuan Zhang, and Xinwei He.
\newblock Levit-unet: Make faster encoders with transformer for medical image
  segmentation.
\newblock {\em arXiv preprint arXiv:2107.08623}, 2021.

\bibitem{yan2020self}
Ke Yan, Jinzheng Cai, Dakai Jin, Shun Miao, Adam~P Harrison, Dazhou Guo, Youbao
  Tang, Jing Xiao, Jingjing Lu, and Le Lu.
\newblock Self-supervised learning of pixel-wise anatomical embeddings in
  radiological images.
\newblock {\em arXiv preprint arXiv:2012.02383}, 2020.

\bibitem{yu2020c2fnas}
Qihang Yu, Dong Yang, Holger Roth, Yutong Bai, Yixiao Zhang, Alan~L Yuille, and
  Daguang Xu.
\newblock C2fnas: Coarse-to-fine neural architecture search for 3d medical
  image segmentation.
\newblock In {\em Proceedings of the IEEE/CVF Conference on Computer Vision and
  Pattern Recognition}, 2020.

\bibitem{zhai2021scaling}
Xiaohua Zhai, Alexander Kolesnikov, Neil Houlsby, and Lucas Beyer.
\newblock Scaling vision transformers.
\newblock {\em arXiv preprint arXiv:2106.04560}, 2021.

\bibitem{zheng2020rethinking}
Sixiao Zheng, Jiachen Lu, Hengshuang Zhao, Xiatian Zhu, Zekun Luo, Yabiao Wang,
  Yanwei Fu, Jianfeng Feng, Tao Xiang, Philip~HS Torr, et~al.
\newblock Rethinking semantic segmentation from a sequence-to-sequence
  perspective with transformers.
\newblock {\em arXiv preprint arXiv:2012.15840}, 2020.

\bibitem{zheng2021rethinking}
Sixiao Zheng, Jiachen Lu, Hengshuang Zhao, Xiatian Zhu, Zekun Luo, Yabiao Wang,
  Yanwei Fu, Jianfeng Feng, Tao Xiang, Philip~HS Torr, et~al.
\newblock Rethinking semantic segmentation from a sequence-to-sequence
  perspective with transformers.
\newblock In {\em Proceedings of the IEEE/CVF Conference on Computer Vision and
  Pattern Recognition}, 2021.

\bibitem{zhou2021nnformer}
Hong-Yu Zhou, Jiansen Guo, Yinghao Zhang, Lequan Yu, Liansheng Wang, and Yizhou
  Yu.
\newblock nnformer: Interleaved transformer for volumetric segmentation.
\newblock {\em arXiv preprint arXiv:2109.03201}, 2021.

\bibitem{zhou2019prior}
Yuyin Zhou, Zhe Li, Song Bai, Chong Wang, Xinlei Chen, Mei Han, Elliot Fishman,
  and Alan~L Yuille.
\newblock Prior-aware neural network for partially-supervised multi-organ
  segmentation.
\newblock In {\em Proceedings of the IEEE/CVF International Conference on
  Computer Vision}, 2019.

\bibitem{zhou2021models}
Zongwei Zhou, Vatsal Sodha, Jiaxuan Pang, Michael~B Gotway, and Jianming Liang.
\newblock Models genesis.
\newblock {\em Medical image analysis}, 67:101840, 2021.

\bibitem{zhu2020rubik}
Jiuwen Zhu, Yuexiang Li, Yifan Hu, Kai Ma, S~Kevin Zhou, and Yefeng Zheng.
\newblock Rubik’s cube+: A self-supervised feature learning framework for 3d
  medical image analysis.
\newblock {\em Medical image analysis}, 64:101746, 2020.

\end{thebibliography}
}
\newpage
\section*{Appendix}
\appendix
\renewcommand{\thesection}{\Alph{section}}
\renewcommand\thefigure{S.\arabic{figure}}
\setcounter{figure}{0}
\renewcommand\thetable{S.\arabic{table}}
\setcounter{table}{0}
\renewcommand\thealgorithm{S.\arabic{algorithm}}
\setcounter{algorithm}{0}
We provide the supplementary materials in the following. In Sec.~\ref{sec:pre_data}, we describe the details of datasets that are used for pre-training from public sources. In Sec.~\ref{sec:preprocess}, we illustrate the preprocessing and implementation details of fine-tuning tasks using BTCV and MSD datasets. In Sec.~\ref{sec:msd_bench}, we present qualitative and quantitative comparisons of segmentation tasks in MRI modality from MSD dataset. The presented results include benchmarks from all top-ranking methods using the MSD test leaderboard. In Sec.~\ref{sec:compl_study}, the model complexity analysis is presented. Finally, we provide pseudocode of Swin UNETR self-supervised pre-training in  Sec.~\ref{sec:pretrain}.

\section{Pre-training Datasets}
\label{sec:pre_data}
In this section, we provide additional information for our pre-training datasets. The proposed Swin UNETR is pre-trained using five collected datasets. The total data cohort contains $5,050$ CT scans of various body region of interests (ROI) such as head, neck, chest, abdomen, and pelvis. LUNA16~\cite{setio2017validation}, TCIA Covid19~\cite{desai2020chest} and LiDC~\cite{armato2011lung} contain $888$, $761$ and $475$ CT scans which composes the chest CT cohort. The HNSCC~\cite{grossberg2018imaging} has $1,287$ CT scans from head and neck squamour cell carcinoma patients. The TCIA Colon dataset~\cite{johnson2008accuracy} comprises the abdomen and pelvis cohort with $1,599$ scans. We split 5\% of each dataset for validation in the pre-training stage. Table~\ref{tab:tab1} summarizes sources of each collected dataset. Overall, the number of training and validation volumes are $4,761$ and $249$, respectively. The Swin UNETR encoder is pre-trained using only unlabeled images, annotations were not utilized from any of theses datasets. We first clip CT image intensities from $-1000$ to $1000$, then normalize to $0$ and $1$. To obtain informative patches of covering anatomies, we crop sub-volumes of $96 \times 96 \times 96$ voxels at foregrounds, and exclude full air (voxel = $0$) patches. In summary, Swin UNETR is pre-trained via a diverse set of human body compositions, and learn a general-purpose representation from different institutes' data that can be leveraged for wide range of fine-tuning tasks.

\section{Preprocessing Pipelines}
\label{sec:preprocess}
We report fine-tuning results on two public benchmarks: BTCV ~\cite{landman2015miccai} and MSD challenge ~\cite{simpson2019large}. BTCV contains 30 CT scans with 13 annotated anatomies and can be formulated as a single multi-organ segmentation task. The MSD contains 10 tasks for multiple organs, from different sources and using different modalities. Details regarding preprocessing these datasets are provided in the subsequent sub-sections of 2.1 and 2.2.

\subsection{BTCV Dataset}
   All CT scans are interpolated into the isotropic voxel spacing of $[1.5 \times 1.5 \times 2.0]~mm$. The multi-organ segmentation problem is formulated as a $13$ class segmentation, which includes large organs such as liver, spleen, kidneys and stomach; vascular tissues of esophagus, aorta, IVC, splenic and portal veins; small anatomies of gallbladder,  pancreas and adrenal glands. Soft tissue window is used for clipping the CT intensities, then normalized to $0$ and $1$ followed by random sampling of $96 \times 96 \times 96$ voxels. Data augmentation of random flip, rotation and intensities shifting are used for training, with probabilities of $0.1$, $0.1$, and $0.5$, respectively.

\begin{figure*}[p]
\centering
\includegraphics[width=\textwidth]{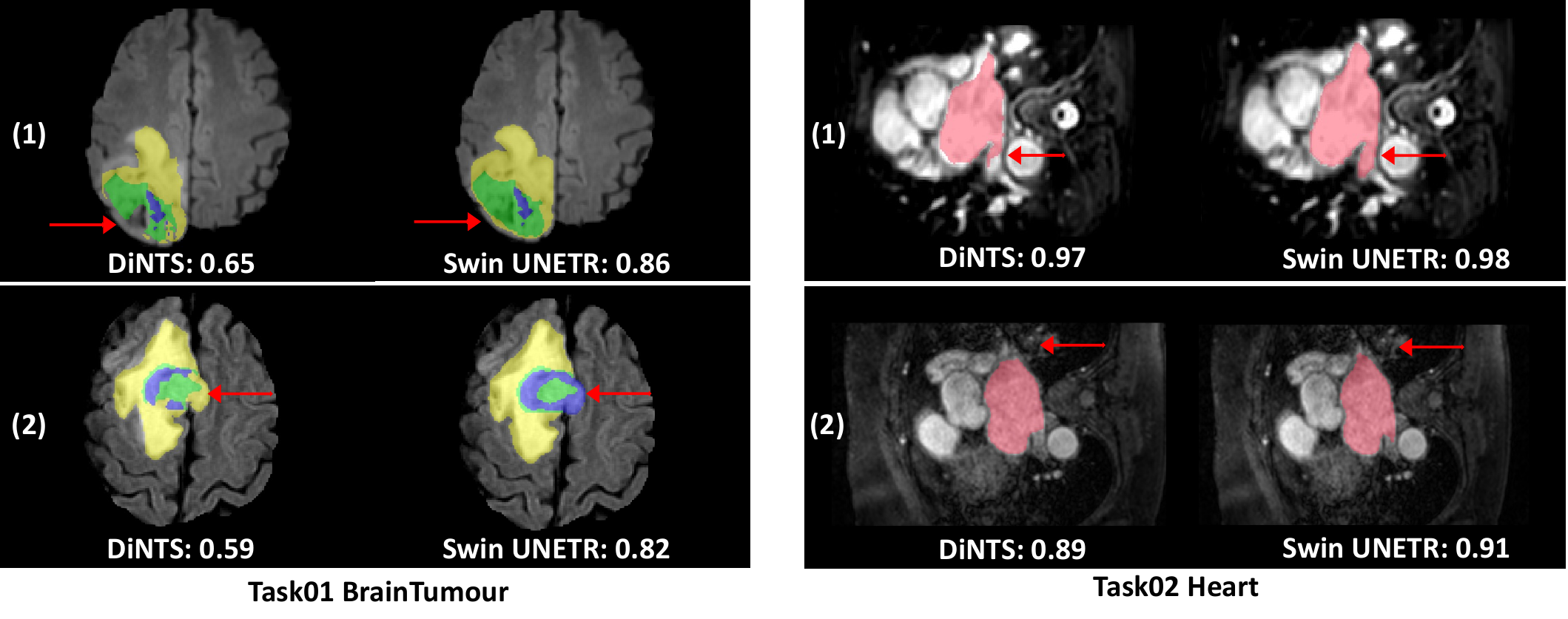}
\includegraphics[width=\textwidth]{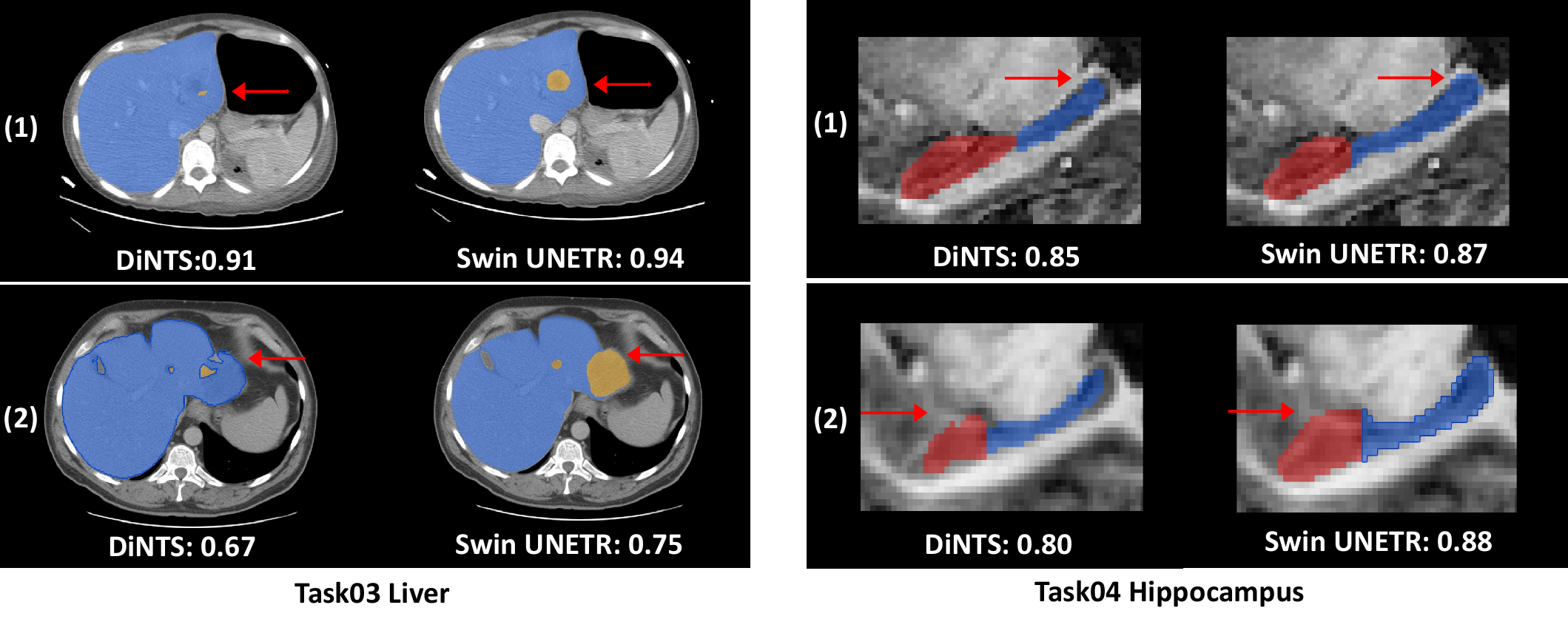}
\includegraphics[width=\textwidth]{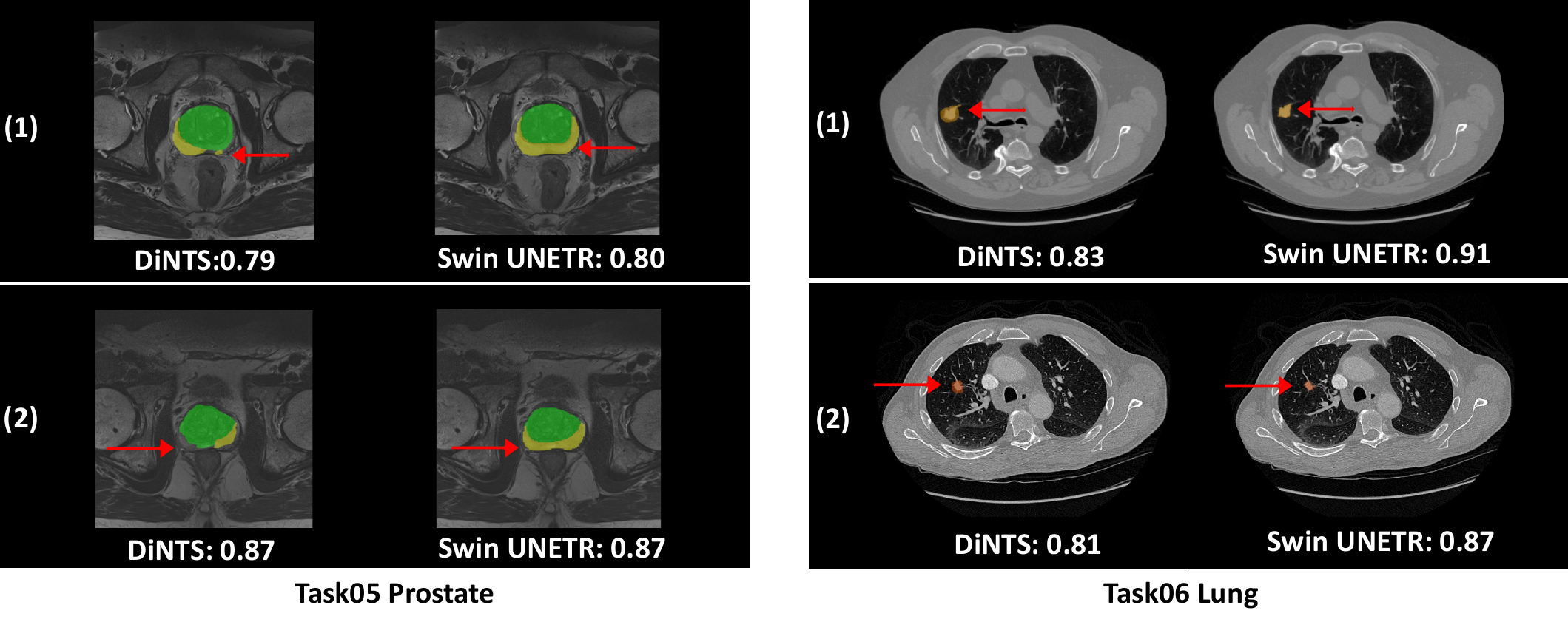}
  \label{fig:fig1_supp}
\end{figure*}
\begin{figure*}[t!]
\centering
\includegraphics[width=\textwidth]{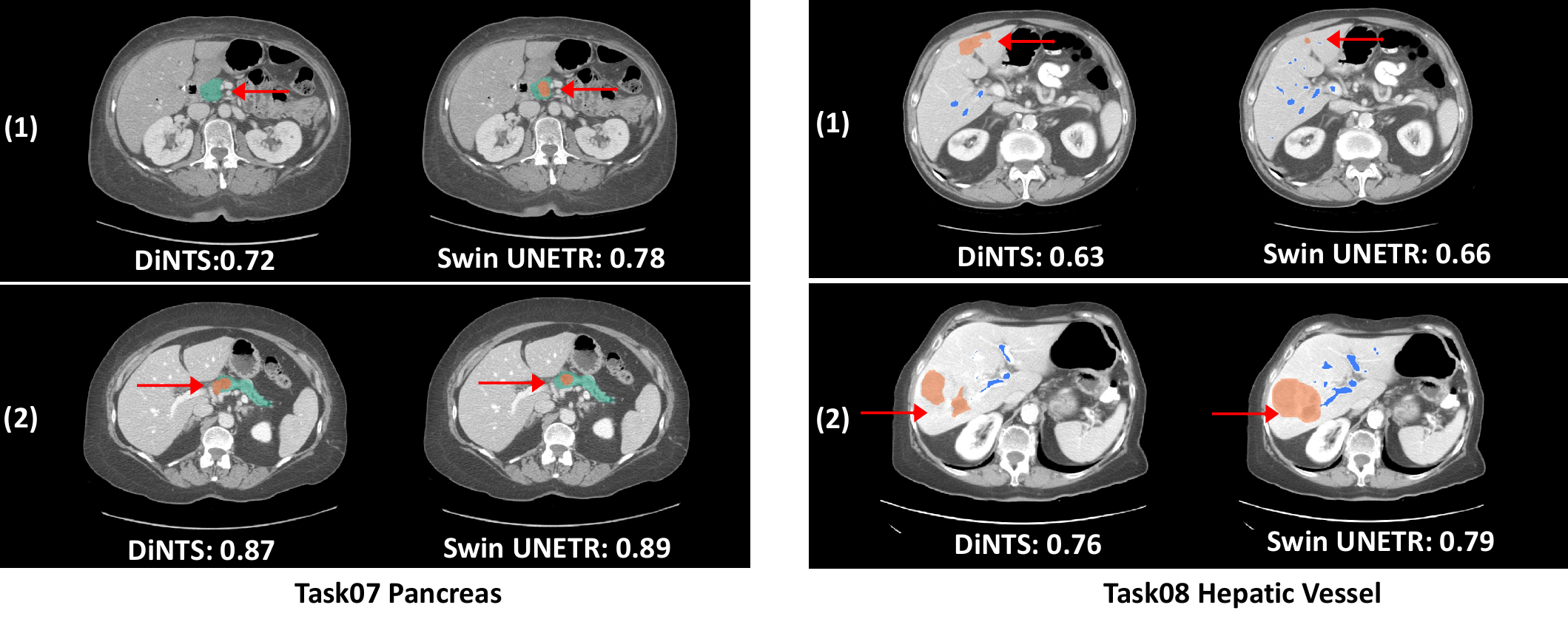}
\includegraphics[width=\textwidth]{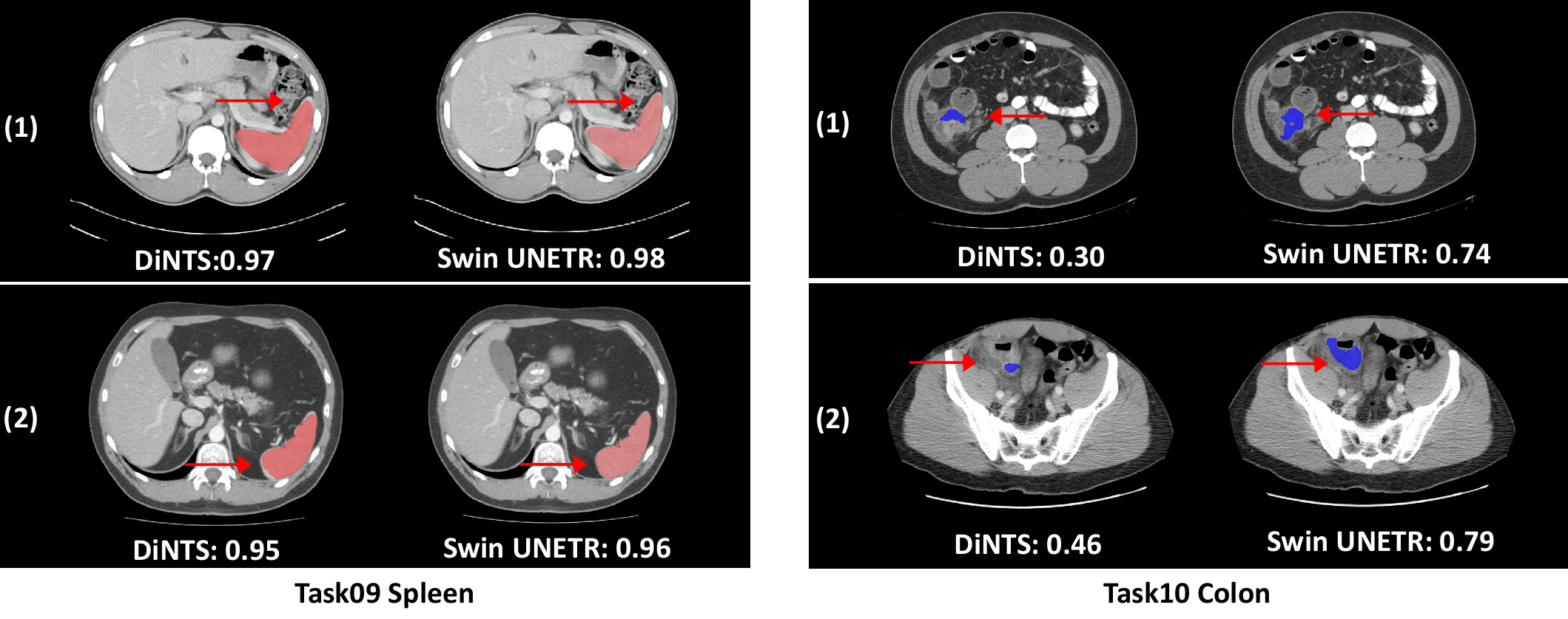}
  \caption{Qualitative visualizations of the proposed Swin UNETR and DiNTS on MSD Tasks}
  \label{fig:fig2_supp}
\end{figure*}

\begin{table*}[!htbp]
 \centering

\resizebox{\textwidth}{!}{
\begin{tabular}{lrrrr}
\hline
Dataset   &Region of Interest & \#Total Samples & Source & Train/Validation \\ 
\hline
LUNA16 \cite{setio2017validation}   & Chest   & 888  & luna16.grand-challenge.org/Data/  & 844/44 \\ 
TCIA Covid19 \cite{desai2020chest} & Chest & 761  & wiki.cancerimagingarchive.net/display/Public/COVID-19 & 723/38 \\
HNSCC \cite{grossberg2018imaging} & Head/Neck & 1287 & wiki.cancerimagingarchive.net/display/Public/HNSCC & 1223/64 \\
TCIA Colon \cite{johnson2008accuracy} & Abdomen/pelvis & 1599 & www.cancerimagingarchive.net/collections/ & 1520/79\\
LiDC \cite{armato2011lung} & Chest & 475 & wiki.cancerimagingarchive.net/display/Public/LIDC-IDRI & 451/24 \\
\hline
\end{tabular}%
}
\caption{Summary of datasets for pre-training, the use of cohorts identifies diversified regions of interest. }
\label{tab:tab1}
\end{table*}

\begin{table*}[t!]
 \centering
\resizebox{\textwidth}{!}{
\begin{tabular}{l|rr|rrr|rrr|rrr|rrr|rr}
\hline
Organ  & \multicolumn{2}{c|}{Task02 Heart} & \multicolumn{6}{c|}{Task04 Hippocampus} & \multicolumn{6}{c|}{Task05 Prostate} & \multicolumn{2}{c}{MRI tasks Avg} \\
\hline
Metric & DSC1 & NSD1  & DSC1  & DSC2  & Avg.  & NSD1 & NSD2 & Avg. & DSC1  & DSC2  & Avg.  & NSD1 & NSD2 & Avg. & DSC & NSD\\ 
\hline
Kim et al~\cite{kim2019scalable}    & 93.11    & 96.44  & 90.11 & 88.72   & 89.42  & 97.77 & \textbf{97.73} & 97.75 & 72.64 & 89.02   & 80.83  & 95.05 & 98.03 & 96.54 & 80.96 & 93.43 \\
Trans VW~\cite{haghighi2021transferable}     & \textbf{93.33}    & 96.51  & \textbf{90.29} & \textbf{88.77} & \textbf{89.53}   & \textbf{97.87}  & 97.67 & \textbf{97.77}  & 73.69 & 88.88 & 81.29   & 95.42  & 98.52 & 96.97 & 81.32 & 93.72\\
C2FNAS\cite{yu2020c2fnas}       & 92.49    & 95.81  & 89.37 & 87.96   & 88.67  & 97.27 & 97.35 & 97.31 & 74.88 & 88.75   & 81.82  & \textbf{98.79} & 95.12 & 96.96 & 81.24 & 93.49\\
Models Gen~\cite{zhou2021models}     & \textbf{93.33}    & 96.51  & \textbf{90.29} & \textbf{88.77}   & \textbf{89.53}  & \textbf{97.87}  & 97.67 & \textbf{97.77} & 73.69 & 88.88   & 81.29  & 95.42 & 98.52 & 96.97 & 81.32 & 93.72\\
nnUNet~\cite{isensee2021nnu}     & 93.30    & \textbf{96.74}  & 90.23 & 88.69   & 89.46  & 97.79 & 97.53 & 97.75 & \textbf{76.59} & \textbf{89.62}   & \textbf{83.11}  & 96.27 & \textbf{98.85} & \textbf{97.56} & 81.74 & 93.91 \\
DiNTS~\cite{he2021dints}     & 92.99    & 96.35  & 89.91 & 88.41   & 89.16  & 97.76 & 97.56 & 97.66 & 75.37 & 89.25   & 82.31  & 95.96 & 98.82 & 97.39 & 81.76 & 94.03\\
\hline
SwinUNETR   & 92.62 & 96.23 & 89.95 & 88.42 & 89.19 & 97.63 & 97.32 & 97.48 & 75.65 & 89.15 & 82.40 & 95.89 & 98.70 & 97.30 & \textbf{82.14} & \textbf{94.66}
\\

\hline

\end{tabular}%
}
\caption{Additional MSD MRI test dataset performance comparison of Dice and NSD. Benchmarks obtained from MSD test leaderboard. Task01 BrainTumuor results are shown in the paper. Note: The results reported for TransVW~\cite{haghighi2021transferable} and Models Genesis~\cite{zhou2021models} are from the official leaderboard for MRI tasks.}
\label{tab:task0604}
\end{table*}

\begin{table}[t!]
 \centering
\resizebox{1\columnwidth}{!}{
\begin{tabular}{lrrr}
\hline
Models   & \#Params (M)  & FLOPs (G) & Inference Time (s) \\
\hline
nnUNet \cite{isensee2021nnu}   & 19.07   & 412.65  & 10.28   \\ 
CoTr \cite{xie2021cotr}   & 46.51   & 399.21  & 19.21   \\ 
TransUNet \cite{chen2021transunet}     & 96.07 & 48.34 & 26.97  \\
ASPP \cite{deeplabv3plus2018}   & 47.92   & 44.87  & 25.47   \\ 
SETR \cite{zheng2020rethinking}     & 86.03 & 43.49 & 24.86 \\
UNETR   & 92.58 & 41.19 & 12.08\\
\bf{SwinUNETR}   & 61.98 & 394.84 & 13.84\\
\hline
\\
\end{tabular}%
}
\caption{ Comparison of number of parameters, FLOPs and averaged inference time for various models in BTCV experiments.}
\label{tab:complexity}
\end{table}

\subsection{MSD Dataset}
The MSD challenge contains $6$ CT and $4$ MRI datasets. We provide additional parameters of pre-processing and augmentation details for each task as follows:

\noindent\textbf{Task01 BrainTumour: } The four modalities MRI images for each subject are formed into $4$ channels input. We convert labels to multiple channels based on tumor classes. which label $1$ is the peritumoral edema, label $2$ is the GD-enhancing tumor, and label $3$ is the necrotic and non-enhancing tumor core. Label $2$ and $3$ are merged to construct tumor core (TC), label $1$, $2$ and $3$ are merged to construct whole tumor (WT), and label $2$ is the enhancing tumor (ET). We crop the sub-volume of $128 \times 128 \times 128$ voxels and use channel-wise nonzero normalization for MRI images. Data augmentation probabilities of $0.5$, $0.1$ and $0.1$ are set for random flips at each axis, intensities scaling and shifting, respectively. 

\noindent\textbf{Task02 Heart: } The heart MRI images are interpolated to the isotropic voxel spacing of $1.0$ $mm$. Channel-wise nonzero normalization is applied to each scan. We sample the training sub-volumes of $96 \times 96 \times 96$ voxels by ratio of positive and negative as 2:1. Augmentation probabilities for random flip, rotation, intensities scaling and shifting are set to $0.5$, $0.1$, $0.2$, $0.5$, respectively.

\noindent\textbf{Task03 Liver: } Each CT scan is interpolated to the isotropic voxel spacing of $1.0$ $mm$. Intensities are scaled to $[-21, 189]$, then normalized to $[0, 1]$. 3D patches of $96 \times 96 \times 96$ voxels are obtained by sampling positive and negative ratio of $1:1$. Data augmentation of random flip, rotation, intensities scaling and shifting are used, for which the probabilities are set to $0.2$, $0.2$, $0.1$, $0.1$, respectively.

\noindent\textbf{Task04 Hippocampus: } Each hippocampus MRI image is interpolated by voxel spacing of $0.2 \times 0.2 \times 0.2$, then applied spatial padding to $96 \times 96 \times 96$ as the input size of Swin UNETR model. Same as other MRI datasets, channel-wise nonzero normalization is used for intensities. Probability of $0.1$ is used for random flip, rotation, intensity scaling \& shifting. 

\noindent\textbf{Task05 Prostate: } We utilize both given modalities for prostate MRI images for each subject as two channels input. Channel-wise nonzero normalization is used. Voxel spacing of $0.5$ and spatial padding of each axis are employed to construct the input size of $96 \times 96 \times 96$. We use random flip, rotation, intensity scaling and shifting with probabilities of $0.5$ as data augmentations. Random affine is applied as additional transformation with scale factor of $[0.3, 0.3, 0.0]$ and rotation range of $[0, 0, pi]$ at each axis.

\noindent\textbf{Task06 Lung: } We interpolate each image to isotropic voxel spacing of $1.0$. Houndsfield unit (HU) range of [-1000, 1000] is used and normalized to $[0, 1]$. Subsequently, training sample are cropped to $96 \times 96 \times 96$ with positive and negative ratio of $2:1$. Augmentation probabilities of $0.5$, $0.3$, $0.1$, $0.1$ are used for random flip, rotation, intensities scaling and shifting. 

\noindent\textbf{Task07 Pancreas: } We clip the intensities to a range of $-87$ to $199$. Patch size of $96 \times 96 \times 96$ is used to sample training data with positive and negative ratio of $1:1$. We set augmentation of random flip, rotation and intensity scaling to probabilities of $0.5$, $0.25$ and $0.5$, respectively.

\noindent\textbf{Task08 HepaticVessel: } To fit the optimal tissue window for hepatic vessel and tumor, we clip each CT image intensities to $[0, 230]$ HU. We apply data augmentation same with Task07 Pancreas for training.

\noindent\textbf{Task09 Spleen: } Spleen CT scans are pre-process with interpolation isotropic voxel spacing of $1.0$ $mm$ on each axis. Soft tissue window of $[-125, 275]$ HU is used for the portal venous phase contrast enhanced CT images. We use the training data augmentation of random flip, intensity scaling \& shifting with probabilities of $0.15$, $0.1$, and $0.1$, respectively.

\noindent\textbf{Task10 Colon: } We use HU range of $[-57, 175]$ for the colon tumor segmentation task and normalized to $0$ and $1$. Next, we sample training sub-volumes by positive and negative ratio of $1:1$. Same as Task07 and Task08, we use random flip, rotation, intensity scaling as augmentation transforms with probabilities of $0.5$, $0.25$ and $0.5$, respectively.

\section{Results}
\label{sec:msd_bench}
\subsection{MSD Qualitative Comparisons}
In this section, we provide extensive segmentation visualization from MSD dataset. In particular, we compare two cases randomly selected from Swin UNETR and DiNTS for each MSD task. As shown in Fig~\ref{fig:fig2_supp}, DiNTS includes the under-segmentation due to lack of parts of labels (Heart, Hippocampus). The missing parts result in a lower Dice score. On BrainTumour, Liver, Pancreas, HepaticVessel and Colon tasks, the comparison indicate that our method achieves better segmentation where the under-segmentation of tumors are observed in DiNTS. For Lung task, the over-segmentation is observed with DiNTS where surrounding tissues are included with label of the lung cancer, while Swin UNETR clearly delineate the boundary. In Heart and Spleen, DiNTS and Swin UNETR have comparable Dice score, yet Swin UNETR performs better segmentation on tissue corner (See Fig~\ref{fig:fig2_supp}). Overall, Swin UNETR achieves better segmentation results and solves the under- and over-segmentation outliers as observed in segmentation via DiNTS.

\subsection{MSD Quantitative Comparisons}
In this section, we provide the quantitative benchmarks of MRI segmentation tasks from MSD dataset. In addition to Task01 BrainTumour, we implement experiment on three remaining MRI dataset including Heart, Hippocampus and Prostate (see Table.~\ref{tab:task0604}). The results are directly obtained from the MSD\footnote{{\url{https://decathlon-10.grand-challenge.org/evaluation/challenge/leaderboard/}}} leaderboard. Regarding MRI benchmark, we achieve much better performance on brain tumor segmentation presented in the paper, with  average Dice improvement of $2\%$ against second best performance. Comparing to models genesis~\cite{zhou2021models}, nnUNet~\cite{isensee2021nnu}, the Swin UNETR shows comparable results on Heart, Hippocampus and Prostate. Overall, we achieve the best average results (Dice of $82.14\%$ and NSD of $94.66\%$) across four MRI datasets, showing Swin UNETR's superiority of medical image segmentation.

\begin{algorithm*}[t!]
\caption{Pytorch Pseudocode of Swin UNETR Self-Supervised Pre-training.}\label{alg:only_one}
  \begin{algorithmic}
\small 
\State \texttt{\textbf{{\color{blue}\# RandRot: transforms of random rotation}}}
\State \texttt{\textbf{\color{blue}\# Cutout: transforms of cutout}}
\State \texttt{\textbf{\color{blue}\# Encoder: swin transformer encoder}}
\State \texttt{\textbf{\color{blue}\# RecHead: reconstruction head}}
\State \texttt{\textbf{\color{blue}\# RotHead: rotation head}}
\State \texttt{\textbf{\color{blue}\# CnHead: contrastive head}}
\State \texttt{\textbf{\color{blue}\# Linpaint: reconstruction loss}}
\State \texttt{\textbf{\color{blue}\# Lrot: rotation loss}}
\State \texttt{\textbf{\color{blue}\# Lcontrast: contrastive loss}}

\State \texttt{\textbf{for x in Loader: {\color{blue}\# minibatch of samples}}}
    \State \hspace{0.4cm} \texttt{\textbf{x1, rot1 = RandRot(x)}}
    \State \hspace{0.4cm} \texttt{\textbf{x2, rot2 = RandRot(x)}}
    \State \hspace{0.4cm} \texttt{\textbf{x1', x2' = Cutout(x1), Cutout(x2)}}
\State \hspace{0.4cm} \texttt{\textbf{z1, z2 = Encoder(x1'), Encoder(x2')}}
    
    \State \hspace{0.4cm} \texttt{\textbf{rec1, rec2 = RecHead(z1), RecHead(z2)}}
    \State \hspace{0.4cm} \texttt{\textbf{contr1, contr2 = CnHead(z1), CnHead(z2)}}
    \State \hspace{0.4cm} \texttt{\textbf{r1, r2 = RotHead(z1), RotHead(z2)}}

    \State \hspace{0.4cm} \texttt{\textbf{rot, r = torch.cat(rot1, rot2), torch.cat(r1, r2)}}
    \State \hspace{0.4cm} \texttt{\textbf{rec, x = torch.cat(rec1, rec2), torch.cat(x1, x2)}}
    \State \hspace{0.4cm} \texttt{\textbf{loss = Linpaint(rec,x) + Lrot(r,rots) + Lcontrast(contr1,contr2)}}
    \State \hspace{0.4cm} \texttt{\textbf{loss.backward() {\color{blue}\# back-propagate}}} \\



\end{algorithmic}
\end{algorithm*}

\section{Model Complexity and Pre-training Time}
\label{sec:compl_study}
In this section, we examine the model complexity along with inference time. In Table.~\ref{tab:complexity}, the number of network paramerts, FLOPs, and averaged inference time of Swin UNETR and baselines on BTCV dataset are presented. We calculate the FLOPs and inference time based on input size of $96 \times 96 \times 96$ used in the BTCV experiments with sliding window approach. Swin UNETR shows moderate size of parameter with $61.98$M, less than transformer-based methods such as TransUNet~\cite{chen2021transunet} of $96.07$M, SETR~\cite{zheng2021rethinking} of $86.03$M, and UNETR~\cite{hatamizadeh2021unetr} of $92.58$M, but larger than 3DUNet (nnUNet)~\cite{isensee2021nnu} of $19.07$M, ASPP~\cite{deeplabv3plus2018} $47.92$M. Our model also shows comparable FLOPs and inference time in terms of 3D approaches such as nnUNet~\cite{isensee2021nnu} and CoTr~\cite{xie2021cotr}. Overall, Swin UNETR outperforms CNN-based and other transformer-based methods while perserves moderate model complexity. Regarding self-supervised pre-training time of Swin UNETR encoder, our approach takes only approximately 6 GPU days. We evaluate pre-training on the 5 collected public datasets with totally 5,050 scans for training and validation, and set maximum training iterations to 45K steps.

\section{Pre-Training Algorithm Details}
\label{sec:pretrain}
In this section, we illustrate the Swin UNETR pre-training details. The Pytorch-like pseudo-code implementation is shown in Algorithm~\ref{alg:only_one}. The Swin UNETR is trained in self-supervised learning paradigm, where we design masked volume inpainting, rotation prediction and contrastive coding as proxy tasks. The self-training aims at improving the quality of representations learnt by large unlabeled data and propagating to smaller fine-tuning dataset. To this end, we leverage multiple transformations for input 3D data, which can exploit inherent context by a mechanism akin to autoencoding and similarity identification. In particular, given an input mini batch data, the transform of random rotation is implemented on each image in the mini batch iteratively. To simultaneously utilize augmentation transformations for contrastive learning, the random rotation of $0^{\circ}$, $90^{\circ}$, $180^{\circ}$, $270^{\circ}$ is applied twice on the same input to generate randomly augmented image pairs of the same image patch. Subsequently, the mini batch data pairs are constructed with the cutout transforms. The drop size of voxels are set to $30\%$ of input sub-volumes. We randomly generate masked ROIs inside image, until the total masked voxels are larger than scheduled number of dropping voxels. Unlike canonical pre-training rules of masked tokens in BERT~\cite{devlin2018bert}, our local transformations to the CT sub-volumes are then arranged to neighbouring tokens. This scheme can construct semantic targets across partitioned tokens, which is critical in medical spatial context. By analogy to Models Genesis~\cite{zhou2021models}, which is CNN-based model consisting expensive convolutional, transposed convolution layers and skip connection between encoder and decoder, our pre-training approach is trained to reconstruct input sub-volumes from the output tokens of the Swin Transformer. Overall, the intuition of modeling inpainting, rotation prediction and contrastive coding is to generalize better representations from aspects of images context, geometry and similarity, respectively.



\end{document}